\documentclass[10pt,journal,compsoc]{IEEEtran}
\usepackage{graphicx}
\usepackage{amsmath}
\usepackage{amssymb}
\usepackage{comment}
\usepackage{multirow}
\usepackage{booktabs}
\usepackage{amsmath,amssymb}
\usepackage{graphicx}
\usepackage{subfigure}
\usepackage{wrapfig}
\usepackage{color}

\newcommand{\parsection}[1]{\noindent\textbf{#1}}

\usepackage[pagebackref=true,breaklinks=true,letterpaper=true,colorlinks,bookmarks=false]{hyperref}

\hypersetup{
    colorlinks=true,
    linkcolor=red,
    filecolor=magenta,      
    urlcolor=magenta,
}

\ifCLASSOPTIONcompsoc
  \usepackage[nocompress]{cite}
\else
  \usepackage{cite}
\fi
\ifCLASSINFOpdf
\else
\fi
\begin{document}
%
\title{Occlusion-Aware Instance Segmentation via BiLayer Network Architectures}
%
%
%
%

\author{Lei Ke,
        Yu-Wing Tai,~\IEEEmembership{Senior Member, IEEE,}
        and~Chi-Keung Tang,~\IEEEmembership{Fellow,~IEEE}
\IEEEcompsocitemizethanks{\IEEEcompsocthanksitem L. Ke and C.-K. Tang are with the Department
of Computer Science and Engineering, The Hong Kong University of Science and Technology. E-mail: \{lkeab, cktang\}@cse.ust.hk. \\
\IEEEcompsocthanksitem Y.-W. Tai is with Kuaishou Technology. E-mail: yuwing@gmail.com.}}

%
%

\markboth{}%
{Shell \MakeLowercase{\textit{et al.}}: Bare Demo of IEEEtran.cls for Computer Society Journals}
%



\IEEEtitleabstractindextext{%
\begin{abstract}
Segmenting highly-overlapping image objects is challenging, because there is typically no distinction between real object contours and occlusion boundaries on images. Unlike previous instance segmentation methods, we model image formation as a composition of two overlapping layers, and propose~\textbf{B}ilayer~\textbf{C}onvolutional~\textbf{Net}work (\textbf{BCNet}), where the top layer detects  occluding objects (occluders) and the bottom layer infers partially occluded instances (occludees).
The explicit modeling of occlusion relationship with bilayer structure naturally decouples the boundaries of both the occluding and occluded instances, and considers the interaction between them during mask regression. 
We investigate the efficacy of bilayer structure using two popular convolutional network designs, namely, Fully Convolutional Network (FCN) and Graph Convolutional Network (GCN). Further, we formulate bilayer decoupling using the vision transformer (ViT), by representing instances in the image as separate learnable occluder and occludee queries. Large and consistent improvements using one/two-stage and query-based object detectors with various backbones and network layer choices validate the generalization ability of bilayer decoupling,
as shown by 
extensive experiments on image instance segmentation benchmarks (COCO, KINS, COCOA) and video instance segmentation benchmarks (YTVIS, OVIS, BDD100K MOTS), 
especially for heavy occlusion cases.
Code and data are available at \url{https://github.com/lkeab/BCNet}.
\end{abstract}


}

\maketitle

\IEEEdisplaynontitleabstractindextext

%
\IEEEpeerreviewmaketitle

\IEEEraisesectionheading{\section{Introduction}\label{sec:introduction}}

\IEEEPARstart{S}{tate}-of-the-art approaches in instance segmentation often follow the Mask R-CNN~\cite{he2017mask} paradigm with the first stage detecting  bounding boxes, followed by  the second stage of segmenting instance masks. Mask R-CNN and its variants~\cite{liu2018path,cai2018cascade,chen2018masklab,huang2019mask,chen2019hybrid} have demonstrated notable performance, and most of the leading approaches in the COCO instance segmentation challenge~\cite{lin2014microsoft} have adopted this pipeline. However, we note that most incremental improvement comes from better backbone architecture designs, with little  attention paid in the instance mask regression after obtaining the ROI (Region-of-Interest) features from object detection. We observe that a lot of segmentation errors are caused by overlapping objects, especially for  object instances belonging to the same class. 
This is because each instance mask is individually regressed, and the regression process implicitly assumes the object in an ROI has almost complete contour, since most objects in the training data in COCO do not exhibit significant occlusions. 

\begin{figure}[!t]
	\centering
	\includegraphics[width=1.0\linewidth]{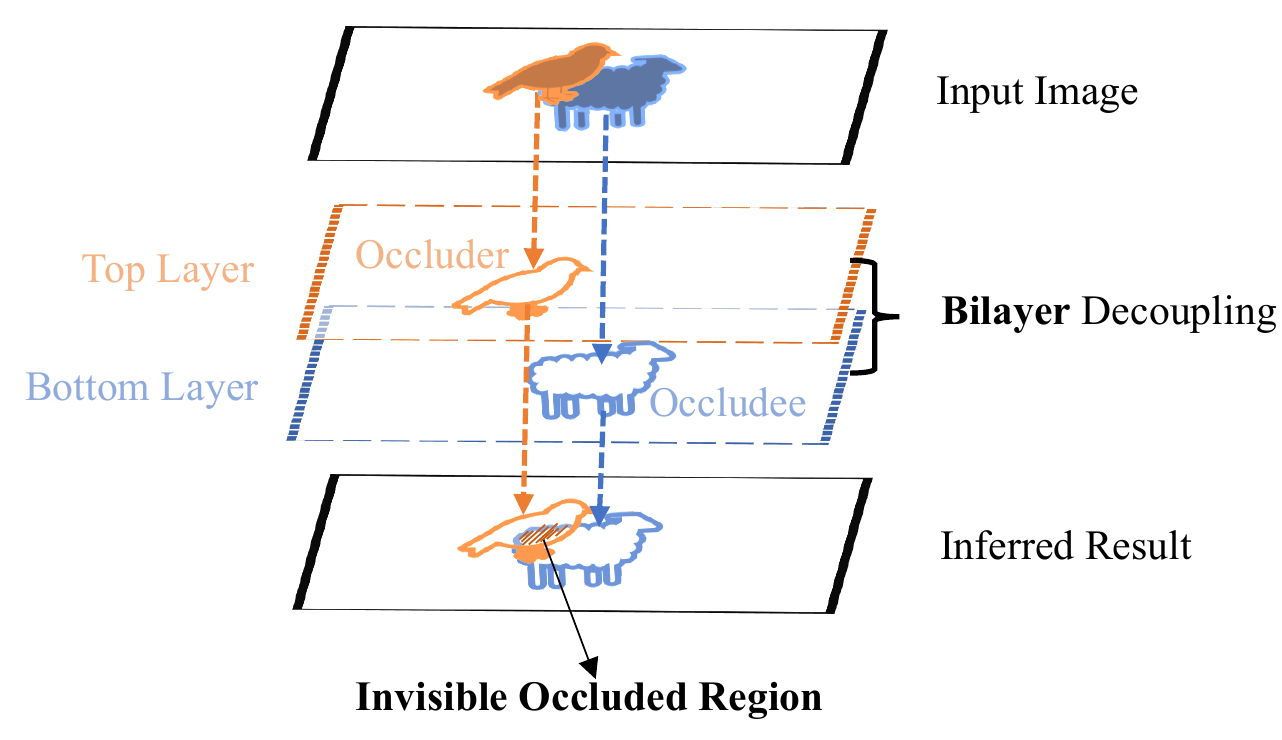}
	\caption{\textbf{Simplified illustration on BCNet's key contribution}. Unlike previous segmentation approaches operating on a single image layer (i.e., directly on the input image), we decouple overlapping objects into~\textit{two image layers}, where the top layer deals with the occluding objects (\textbf{occluder}) and the bottom layer for \textbf{occludee} (which is also referred to as target object in other methods as they do not explicitly consider the occluder). The overlapping parts of the two image layers indicate the invisible region of the occludee, which is explicitly modeled by our occlusion-aware BCNet framework.}
	\label{fig:teaser1}
	\vspace{-0.3cm}
\end{figure}

\begin{figure*}[!t]
	\centering
    \includegraphics[width=1.0\linewidth]{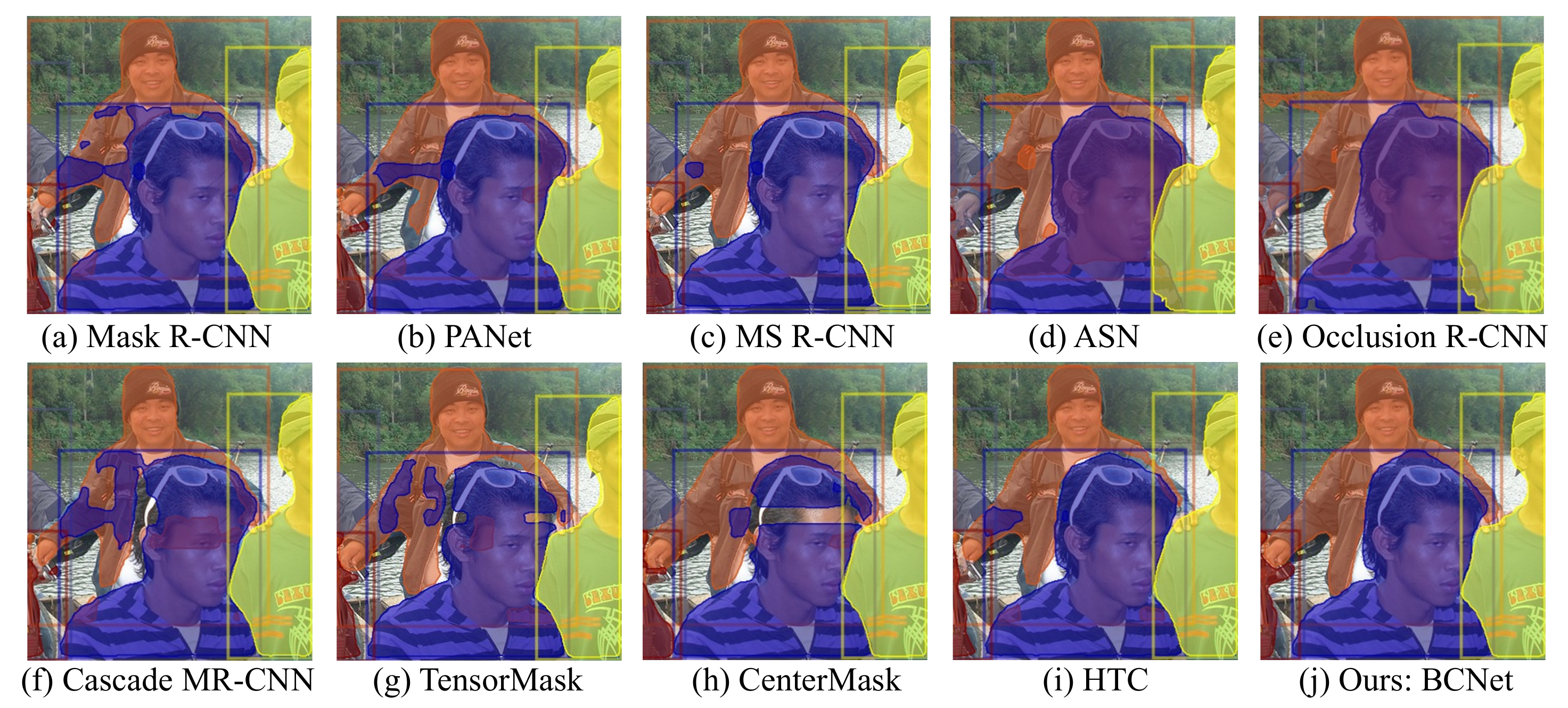}
	\vspace{-0.35in}
	\caption{Instance Segmentation on \textbf{COCO}~\cite{lin2014microsoft} validation set by a) Mask R-CNN~\cite{he2017mask}, b) PANet~\cite{liu2018path}, c) Mask Scoring R-CNN~\cite{huang2019mask}, d) ASN~\cite{qi2019amodal}, e) Occlusion R-CNN (ORCNN)~\cite{follmann2019learning}, f) Cascade Mask R-CNN~\cite{cai2018cascade}, g) TensorMask~\cite{chen2019tensormask}, h) CenterMask~\cite{lee2019centermask}, i) HTC~\cite{chen2019hybrid} and j) Our BCNet. Note that d) and e) are specially designed for amodal/occlusion mask prediction. In this example, the bounding box is given to compare the quality of different regressed instance masks.}
    \label{fig:fig1}
    \vspace{-0.2in}
\end{figure*}

We propose the Bilayer Convolutional Network (BCNet) with its core contribution illustrated in Figure~\ref{fig:teaser1}. BCNet simultaneously regresses both occluding region (occluder) and partially occluded object (occludee) after ROI extraction, which  groups the pixels belonging to the occluding region and treat them equally as the pixels of the occluded object but in \textit{two separate image layers}, and thus naturally decouples the boundaries for both objects 
and considers the interaction between them during the mask regression stage.  

\begin{figure*}[!t]
	\centering
	\includegraphics[width=1.0\linewidth]{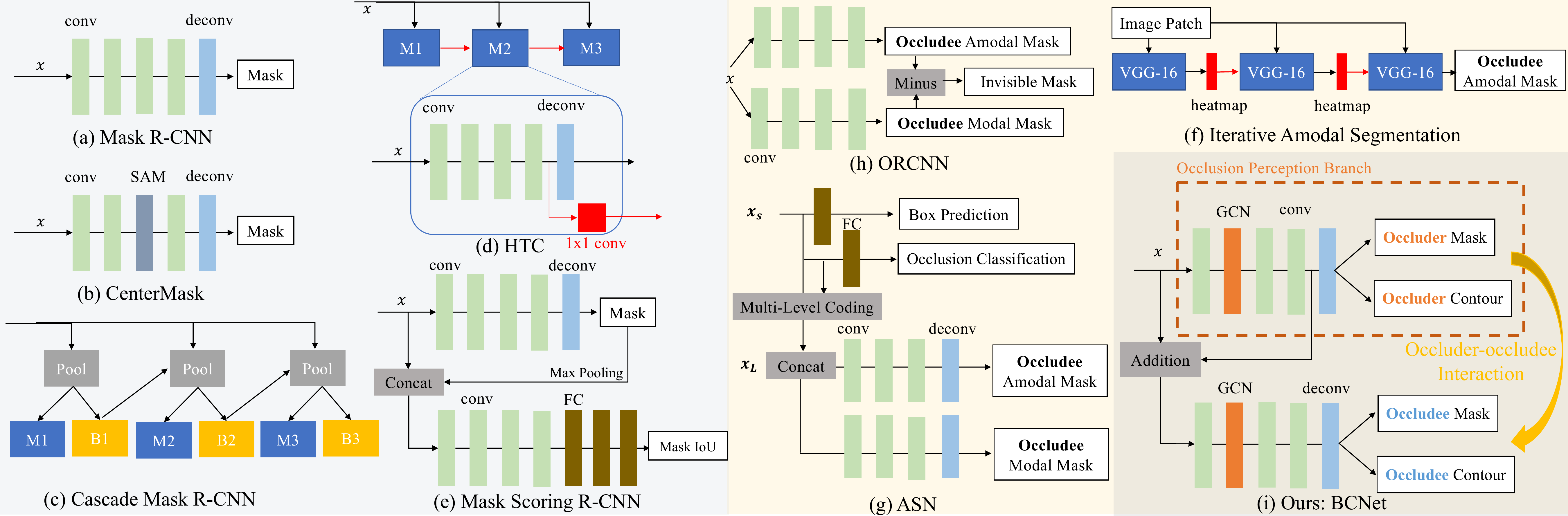}
	\vspace{-0.25in}
	\caption{A brief comparison of \textbf{mask head architectures}: a) Mask R-CNN~\cite{he2017mask}, b) CenterMask~\cite{lee2019centermask}, c) Cascade Mask R-CNN~\cite{cai2018cascade}, d) HTC~\cite{chen2019hybrid}, e) Mask Scoring R-CNN~\cite{huang2019mask}, f) Iterative Amodal Segmentation~\cite{li2016amodal}, g) ASN~\cite{qi2019amodal}, h) ORCNN~\cite{follmann2019learning}, where f), g) and h) are specially designed for amodal/occlusion mask prediction, i) Ours: BCNet. The input $\mathbf{x}$ denotes CNN feature after ROI extraction. {\tt Conv} is convolution layer with $3\times3$ kernel, {\tt FC} is the fully connected layer, {\tt SAM} is the spatial attention module. B$_t$ and M$_t$ respectively denote box and mask head at $t$-th stage. Unlike previous occlusion-aware mask heads, which only regress both modal and amodal masks from the occludee, our BCNet has a~\textit{bilayer GCN structure} and considers the \textbf{interactions between the top ``occluder" and bottom ``occludee"} in the same ROI. The \textbf{occlusion perception branch} explicitly models the occluding object by performing joint mask and contour predictions, and distills essential occlusion information for the second graph layer to segment target object (``occludee''). }
	\label{fig:motivation}
	\vspace{-0.2in}
\end{figure*}

Previous approaches resolve the mask conflict between neighboring objects through  non-maximum suppression or additional post-processing~\cite{liu2016multi,dai2016instance,li2016iterative,krahenbuhl2011efficient,hariharan2015hypercolumns}. Consequently, their results are over-smooth along boundaries or exhibit small gaps between neighboring objects. Furthermore, since the receptive field in the ROI observes multiple objects that belong to the same class, when the occluding regions were included as part of the occluded object, traditional mask head design falls short of resolving such conflict, leaving a large portion of error as shown in Figure~\ref{fig:fig1}.
We  compare BCNet with recent amodal segmentation methods~\cite{qi2019amodal,follmann2019learning}, which predict complete object masks, including the occluded region.
However, these amodal methods only regress single occluded target in the ROI, thus lacking  occluder-occludee interaction reasoning, making their specially designed decoupling structure suffer when handling mask conflict between highly-overlapping objects.
Correspondingly, 
Figure~\ref{fig:motivation} compares the architecture of our BCNet with previous mask head designs~\cite{he2017mask,liu2018path,huang2019mask,chen2019hybrid,lee2019centermask,cai2018cascade,qi2019amodal,follmann2019learning}. 

A preliminary version of BCNet appears in~\cite{ke2021bcnet}. Our BCNet consists of two GCN layers with a cascaded structure, each respectively regresses the mask and boundaries of the occluding and partially occluded objects. We utilize GCN in our implementation because GCN can consider non-local relationship between pixels, allowing for propagating information across pixels despite the presence of occluding regions.
The explicit bilayer occluder-occludee relational modeling within the same ROI also makes our final segmentation results more explainable than previous methods. 
We also experiment BCNet with pure FCN layers, and find that the bilayer structure still generalizes well, despite achieving inferior performance comparing to bilayer GCN.
For object detector, we use the FCOS~\cite{tian2019fcos} owing to its efficient memory and running time, while noting that other state-of-the-art object detectors can also be used as demonstrated in our experiments. 

Besides the aforementioned standard CNN and GCN architecture in the preliminary work~\cite{ke2021bcnet}, we further summarize the extensions as: 1) We implement BCNet using the emerging vision transformer (ViT)~\cite{vit} for instance segmentation. 2) We perform extensive quantitative and qualitative analysis for the transformer-based BCNet, which achieves 44.6 mask AP on COCO by using R50-FPN. 3) We further apply BCNet to three complicated video instance segmentation benchmarks and obtain consistent improvement.

Our transformer-based BCNet explicitly decouples the instance queries by representing image objects into two individual groups, one  representing the occluded objects (occludees), while the other for the corresponding occluding objects (occluders). Instead of using a single transformer decoder~\cite{cheng2021mask2former}, we design a bilayer transformer decoder with a cascaded structure, where the first transformer decoder distills occluder information, which is then injected into the second transformer decoder for occludee mask prediction. In doing so, both instance queries and transformer decoders can perceive the occluder-occludee relations, contributing to the first occlusion-aware transformer structure.

Since our paper focuses on occlusion handling in instance segmentation, in addition to the original COCO evaluation, we extract a subset of COCO dataset containing both occluding objects and partially occluded objects to evaluate the robustness of our approach in comparison with other 
instance segmentation methods in occlusion handling.
In this paper, we also contribute a large-scale occlusion-aware instance segmentation dataset SOD with ground-truth, complete object contours for {\em both} occluding and partially occluded objects. 
Extensive experiments show that our approach outperforms state-of-the-art methods in both the modal and amodal instance segmentation tasks.

\section{Related Work}\label{sec:related_work}
\parsection{Image Instance Segmentation}
Two stage instance segmentation methods~\cite{li2017fully,he2017mask,liu2018path,chen2018masklab,cai2018cascade,chen2019hybrid,chen2019tensormask} achieve state-of-the-art performance by first detecting bounding boxes and then performing segmentation in each ROI region.
FCIS~\cite{li2017fully} introduces the position-sensitive score maps within instance proposals for mask segmentation. 
Mask R-CNN~\cite{he2017mask} extends Faster R-CNN~\cite{ren2015faster} with a FCN branch to segment objects in the detected box.
PANet~\cite{liu2018path} further integrates multi-level feature of FPN to enhance feature representation.
MS R-CNN~\cite{huang2019mask} mitigates the misalignment between mask quality and score.
CenterMask~\cite{lee2019centermask} is built upon the anchor free detector FCOS~\cite{tian2019fcos} with a SAG-Mask branch.
In contrast, our BCNet is a \textit{bilayer} mask prediction network for addressing the issues of heavy occlusion and overlapping objects in two-stage instance segmentation. Experiments validate that our approach leads to significant performance gain on {\em overall} instance segmentation performance not limited to heavily occluded cases. 

One-stage instance segmentation methods remove the bounding box detection and feature re-pooling steps.
AdaptIS~\cite{sofiiuk2019adaptis} produces masks for objects located on point proposals.
PolarMask~\cite{xie2019polarmask} models instance masks in  polar coordinates by instance center classification and dense distance regression.
YOLOACT~\cite{bolya2019yolact} introduces prototype masks with per-instance coefficients. 
SOLO~\cite{wang2019solo} applies the ``instance categories'' concept to directly output instance masks based on  location and size.
Grouping-based approaches~\cite{kirillov2017instancecut,arnab2017pixelwise,liu2017sgn,liu2018affinity,bai2017deep,kong2018recurrent} regard segmentation as a bottom-up grouping task by first producing pixel-wise predictions followed by grouping object instances in the post-processing stage.
There are also some GCN-based segmentation works~\cite{li2020self,lu2020video,pang2019towards}, however, they mainly focus on the general human parsing and semantic segmentation tasks~\cite{li2022deep} without occlusion-aware modeling.

\parsection{Transfomer-based Instance Segmentation} Inspired by DETR~\cite{carion2020end}, transformer-based instance segmentation methods~\cite{QueryInst,dong2021solq,wang2020end,guo2021sotr,hu2021ISTR} regard segmentation as set prediction.
These methods represent the interested objects using instance queries, and jointly perform class, bounding box and mask predictions. QueryInst~\cite{QueryInst} adopts dynamic mask heads with mask information flow. Mask Transfiner~\cite{transfiner,vmt} produces high-quality instance segmentation by taking detected incoherent points as input queries and employing efficient quadtree transformer. Mask2Former~\cite{cheng2021mask2former} designs a masked cross-attention decoder to constrain the attention regions in~\cite{cheng2021maskformer}, while~\cite{wang2022learning} further boosts the query-based models by discriminative learning. Unlike these methods using a shared decoder, our transformer-based BCNet has a bilayer transformer structure with both occluder and occludee decoders. Each transformer decoder deals with the corresponding set of queries, and then communicates through a residue connection.

\parsection{Occlusion Handling} 
Methods for occlusion handling have been proposed~\cite{sun2005symmetric,winn2006layout,gao2011segmentation,chen2015parsing,yang2011layered,hsiao2014occlusion,gao2011segmentation,zhu2017semantic,yan2019visualizing}.
Ghiasi \textit{et al}.~\cite{ghiasi2014parsing} model occlusion by learning deformable models for human pose estimation while~\cite{Ke_2020_ECCV} reconstructs dense 3D shape for vehicle pose.
Tighe \textit{et al}.~\cite{tighe2014scene} build a histogram to predict occlusion overlap scores between two classes for inferring occlusion order in the scene parsing task.
Chen \textit{et al}.~\cite{chen2015multi} handle occlusion by incorporating category specific reasoning and exemplar-based shape prediction for instance segmentation.
For pedestrian occlusion, bi-box regression is proposed in~\cite{zhou2018bi} for both full body and visible part estimation, while repulsion loss~\cite{wang2018repulsion} and aggregation loss~\cite{zhang2018occlusion} are to improve the detection accuracy.
SeGAN~\cite{ehsani2018segan} learns occlusion patterns by segmenting and generating the invisible part of an object.
OCFusion~\cite{lazarow2019learning} uses an additional branch to model instances fusion process for replacing detection confidence in panoptic segmentation.
A self-supervised scene de-occlusion method is proposed in~\cite{zhan2020self} to complete the mask and content for the invisible object parts. VOIN~\cite{ke2021voin} learns to inpaint the occluded video object using occlusion-aware shape and flow completion.

Compared to these methods, our BCNet  tackles occlusion by explicitly modeling occlusion patterns in shape and appearance. This equips the segmentation model  with strong occlusion perception and reasoning capability. Our bi-layer approach can be smoothly integrated into state-of-the-art segmentation framework for end-to-end training.

\parsection{Amodal Instance Segmentation} 
Different from traditional segmentation which only focuses on visible regions, amodal instance segmentation can predict the occluded parts of object instances. Li and Malik~\cite{li2016amodal} first propose a method by extending~\cite{li2016iterative}, which iteratively enlarges the modal bounding box following the direction of high heatmap values and synthetically adds occlusion. Zhu~\textit{et al}.~\cite{zhu2017semantic} propose a COCO amodal dataset with 5000 images from the original COCO and use AmodalMask as a baseline, which is SharpMask~\cite{pinheiro2016learning} trained on amodal ground truth. COCOA~\textit{cls}~\cite{follmann2019learning} augments this dataset by assigning class-labels to the objects while SAIL-VOS dataset  in~\cite{hu2019sail} is targeted for video object segmentation. In autonomous driving, Qi~\textit{et al.}~\cite{qi2019amodal} establish the large-scale KITTI~\cite{geiger2012we} InStance segmentation dataset (KINS) and present ASN to improve  amodal segmentation performance.

Comparing to  most of the amodal and occlusion reasoning methods which regress single occluded object boundary directly on the input (single-layered) image, our BCNet decouples overlapping objects in the same ROI into two disjoint graph layers by predicting the complete object segments (Figure~\ref{fig:teaser1}), 
where the occludee is segmented under the guidance from the shape and location of the occluder.

\section{Occlusion-Aware Instance Segmentation}\label{sec:method}

\begin{figure*}[!t]
	\centering
	\includegraphics[width=1.0\linewidth]{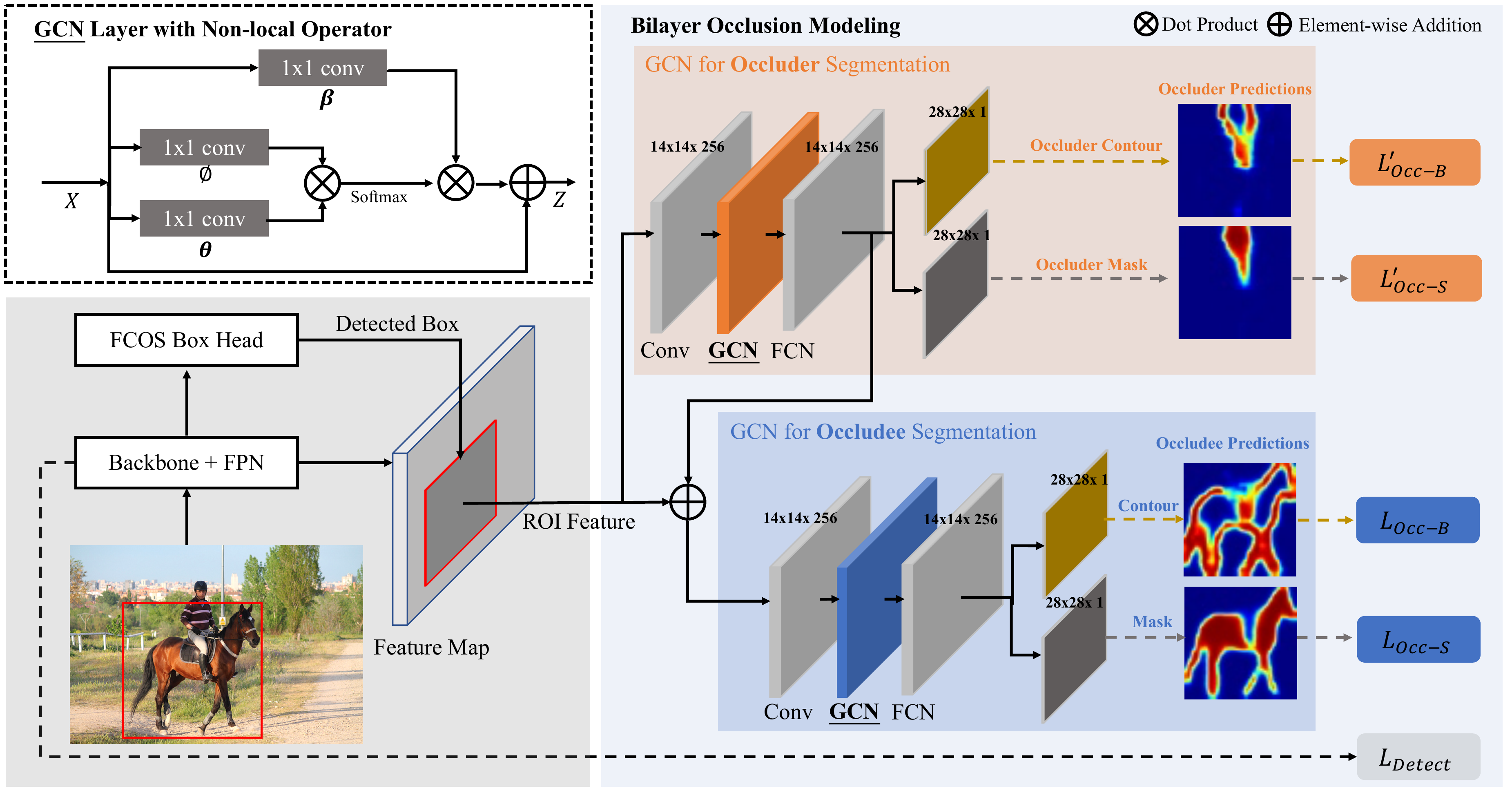}
	\vspace{-0.25in}
	\caption{Architecture of our BCNet for \textbf{GCN-based} Instance Segmentation with bilayer occluder-occludee relational modeling, which consists of three modules; (1)~Backbone~\cite{he2016deep} with FPN for feature extraction from input image; (2)~Detection branch~\cite{tian2019fcos} for predicting instance proposals; (3)~BCNet with bilayer GCN structure for mask prediction. For cropped ROI feature, the first GCN explicitly models occluding regions (occluder) by simultaneously detecting occlusion contours and masks, which distills essential shape and position information to guide the second GCN in mask prediction for the occludee. We utilize the non-local operator~\cite{wang2018non, wang2018videos} detailed in Section~\ref{sec:bilayer_gcn} to implement the GCN layer. Visualization results are resized to squares.}
	\label{fig:framework}
	\vspace{-0.2in}
\end{figure*}

We first describe the explicit occluder-occludee modeling of our proposed Bilayer Convolutional Network (BCNet) in Section~\ref{sec:bilayer}, and then give an overview to the overall bilayer GCN-based instance segmentation framework in Section~\ref{sec:bilayer_gcn}. Based on the principle of bilayer decoupling, we further design a bilayer transformer-based on Mask2Former~\cite{cheng2021mask2former} for occlusion-aware instance segmentation in Section~\ref{sec:bilayer_trans}. Finally, we specify the objective functions for the whole network optimization, and provide details of training and inference process.

BCNet is motivated by images with heavy occlusion, where multiple overlapping objects in the same bounding box may result in confusing instance contours from both real objects and occlusion boundaries. 
The mask head design of Mask R-CNN and its variants~\cite{huang2019mask,chen2019hybrid,cai2018cascade,qi2019amodal,follmann2019learning} in Figure~\ref{fig:motivation} directly regresses the occludee with a fully convolutional network, which neglects both the occluding instances and the overlapping relations between objects. 
To mitigate this limitation, BCNet extends existing two stage instance segmentation methods, by adding an occlusion perception branch parallel to the traditional target prediction pipeline. Thus, the interactions between objects within the ROI region can be well considered  during the mask regression stage.

To obtain occlusion relations among image objects, for amodal instance segmentation, such as KINS~\cite{qi2019amodal} and COCOA~\cite{zhu2017semantic}, ground truth for occluder and occludee is extracted from their annotated object depth/occlusion order. For conventional instance segmentation with no occlusion labeling, such as COCO~\cite{lin2014microsoft}, we simply regard the occludee as the target object inside the bounding box, while the occluder as the union of remaining objects inside the same bounding box with overlapping relation to the target object.

\subsection{Bilayer Occluder-Occludee  Modeling}
\label{sec:bilayer}
\parsection{Bilayer GCN Structure for Instance Segmentation} 
Recently, Graph Convolutional Network (GCN)~\cite{kipf2017semi} has been  adopted to model long-range relationships in images~\cite{chen2019graph,zhang2019dual,li2018beyond} and videos~\cite{wang2018videos}. 
Given highly-overlapping objects, pixels belonging to the same partially occluded object may be separated into disjoint subregions by the occluder.
Thus, we adopt GCN as our basic block due to its non-local property~\cite{wang2018non}, where each graph node represents a single pixel on the feature map.
To explicitly model the occluding region, we further extend the single GCN block to the bilayer GCN structure as shown in Figure~\ref{fig:framework}, which constructs two orthogonal graphs in a single general framework. 

Following~\cite{wang2018videos}, given an adjacency graph~$\mathcal{G=\langle\mathcal{V}, \mathcal{E}}\rangle$ with edges~$\mathcal{E}$ among nodes~$\mathcal{V}$, we represent the graph convolution operation as,
\begin{equation}
\mathbf{Z} = \mathbf{\sigma }(\mathbf{A}\mathbf{X}\mathbf{W}_g) + \mathbf{X},
\end{equation} 
where~$\mathbf{X} \in \mathbb{R}^{N\times K}$ is the input feature, $N=H\times W$ is the number of pixel grids within the ROI region and $K$ is the feature dimension for each node, $\mathbf{A} \in \mathbb{R}^{N\times N}$ is the adjacency matrix for defining neighboring relations of graph nodes by feature similarities, and $\mathbf{W}_g \in \mathbb{R}^{K\times K'}$ is the learnable weight matrix for the output transform, where~$K'=K$ in our case. The output feature $\mathbf{Z} \in \mathbb{R}^{N\times K'}$ consists of the updated node feature by global information propagation within the whole graph layer, which is obtained after non-linear functions~$\mathbf{\sigma }(\cdot)$ including layer normalization~\cite{ba2016layer} and ReLU functions. We add a residual connection after the GCN layer.

To construct the adjacency matrix~$\mathbf{A}$, we define the pairwise similarity between every two graph nodes $\mathbf{x}_i, \mathbf{x}_j$ by dot product similarity as,
\begin{equation}
\mathbf{A}_{ij} = \mathit{softmax}(F(\mathbf{x}_i, \mathbf{x}_j)), 
\end{equation}
\begin{equation}
F(\mathbf{x}_i,\mathbf{x}_j) = \theta(\mathbf{x}_i)^{T}\phi(\mathbf{x}_j),
\end{equation}
where $\theta$ and $\phi$ are two trainable transformation function implemented by $1\times1$ convolution as shown in the non-local operator part of Figure~\ref{fig:framework}, so that high confidence edge between two nodes corresponds to larger feature similarity.

In our bilayer GCN structure, we further define $\mathcal{G}^i$ to indicate the $i$th graph, $X_{roi}$ for the input ROI feature and~$\mathbf{W}_f$ for weights in FCN layers. The pertinent equations are:
\begin{equation}
\mathbf{Z}^1 = \mathbf{\sigma }(\mathbf{A}^1\mathbf{X}_f\mathbf{W}_g^{1}) + \mathbf{X}_{f}, 
\end{equation} 
\begin{equation}
\mathbf{X}_f = \mathbf{Z}^0\mathbf{W}_{f}^0 + \mathbf{X}_{roi}, 
\end{equation}
\begin{equation}
\mathbf{Z}^0 = \mathbf{\sigma}(\mathbf{A}^0\mathbf{X}_{roi}\mathbf{W}_{g}^0)+\mathbf{X}_{roi}.
\end{equation} 
For connecting the two GCN blocks, the output feature $\mathbf{Z}^0$ of the occluder from the first GCN is directly added to~$\mathbf{X}_{roi}$ to obtain the fused~\textit{occlusion-aware} feature~$\mathbf{X}_f$, which is the input for the second GCN layer to output~$\mathbf{Z}^1$ for occludee mask prediction.

Compared to previous class-agnostic mask head with single layer structure, where there is only binary label (foreground/background) per pixel,  the bilayer GCN additionally constructs a new semantic graph space for~\textit{occluding region}. Thus a pixel node in overlapping areas in ROI can concurrently correspond to two different states in bilayer graph. 
While other choices may exist, we believe modeling GCN as a dual-layered structure as shown in Figure~\ref{fig:framework} is a natural choice for handling occlusion.

\parsection{Occluder-occludee Modeling}
We explicitly model occlusion patterns by detecting both contours and masks for the occluders using the first GCN layer.
Since the second GCN layer jointly predicts contours for the occludee, the overlap between the two layers can be directly identified as occlusion boundary which can thus be distinguished from  real object contour (e.g., the occluder and occludee prediction on the rightmost of Figure~\ref{fig:framework}).
The rationale behind this design is that such irregular occlusion boundary unrelated to the occludee is confusing, which in turn provides essential cues for decoupling occlusion relations. Besides, accurate boundary localization  explicitly contributes to segmentation mask prediction.

The module for occluder modeling is designed in a simple yet effective way: one 3$\times$3 convolutional layer followed by one GCN layer and one FCN layer. Then we feed the output to the up-sampling layer and one 1$\times$1 convolutional layer to obtain one channel feature map for joint boundary and mask predictions.
The boundary detection for occluder is trained with loss~$\mathcal{L^\prime}_{\text{Occ-B}}$:
\begin{equation}
\mathcal{L^\prime}_\text{Occ-B} = \mathcal{L}_{\text{BCE}}(W_B\mathcal{F}_{occ}(\mathbf{X}_{roi}), \mathcal{GT}_{B}),
\label{eq:eq4}
\end{equation}
where $\mathcal{L}_{\text{BCE}}$ denotes the binary cross-entropy loss, $\mathcal{F}_{occ}$ denotes the nonlinear transformation function of the occlusion modeling module, $W_{B}$ is the boundary predictor weight, $\mathbf{X}_{roi}$ is the cropped FPN feature map given by RoIAlign operation for the target region, and $\mathcal{GT}_{B}$ is the off-the-shelf occluder boundary that can be readily computed from mask annotations.

For occluder mask prediction, it utilizes the shared feature $\mathcal{F}_{occ}(\mathbf{X}_{roi})$, which is jointly optimized by boundary prediction. The segmentation loss $\mathcal{L^\prime}_{\text{Occ-S}}$ for occluder modeling is designed as
\begin{equation}
\mathcal{L^\prime}_{\text{Occ-S}} = \mathcal{L}_{\text{BCE}}(W_S\mathcal{F}_{occ}(\mathbf{X}_{roi}), \mathcal{GT}_{S}),
\label{eq:eq5}
\end{equation}
where $W_S$ denotes the trainable weight of segmentation mask predictor by $1\times1$ convolutional layer, and $\mathcal{GT}_{S}$ is the mask annotations for the occluder.

\begin{figure*}[!t]
	\centering
	\includegraphics[width=1.0\linewidth]{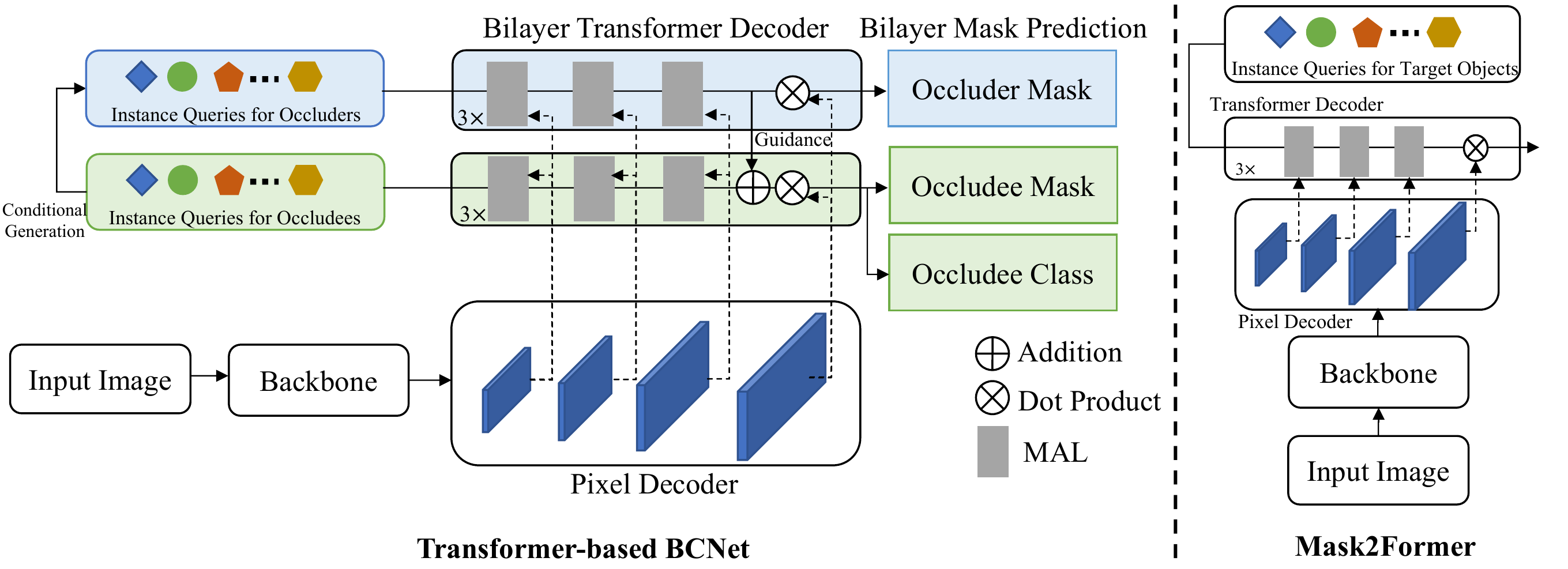}
	\vspace{-0.25in}
	\caption{\textbf{Left}: Architecture of our \textbf{transformer-based} BCNet built on~\cite{cheng2021mask2former} with bilayer transformer decoder. \textbf{Right}: Architecture of Mask2Former~\cite{cheng2021mask2former} for instance segmentation. Instead of adopting a single transformer decoder and only one set of instance queries, our bilayer transformer decoder models occluder-occludee relations by processing occluder and occludee queries in a cascaded manner. In the latter stage of the first transformer decoder, the learned shape and texture information of the occluder is injected to the second decoder to guide the target instance (occludee) segmentation by residue connection. MAL denotes the Masked cross-Attention Layer in ~\cite{cheng2021mask2former}. Pixel decoder constructs a multi-scale feature pyramid from the original image for feeding  into the transformer decoder.}
	\label{fig:framework_trans}
	\vspace{-0.2in}
\end{figure*}

\subsection{Bilayer GCN-based Instance Segmentation}
\label{sec:bilayer_gcn}
Figure~\ref{fig:framework} gives the overall~\textbf{architecture} of BCNet for addressing occlusion in instance segmentation.
Following typical models~\cite{he2017mask,lee2019centermask} for instance segmentation, our model has  three parts:
(1) Backbone~\cite{he2016deep} with FPN~\cite{lin2017feature} for ROI feature extraction;
(2) Object detection head in charge of predicting bounding boxes
as instance proposals. We employ FCOS~\cite{tian2019fcos} as the object detector owing to its anchor-free efficiency though our method is flexible and can deploy any existing fully supervised object detectors~\cite{ren2015faster,redmon2016you,lin2017focal};
(3) The \textit{occlusion-aware mask head}, BCNet, uses bilayer GCN structure for decoupling overlapping relations and segments the instance proposals obtained from the object detection branch.
BCNet reformulates the traditional class-agnostic segmentation as two complementary tasks: occluder modeling using the first GCN and occludee prediction with the second GCN, where the auxiliary predictions from the first GCN provide rich occlusion cues, such as shape and positions of occluding regions, to guide target (occludee) object segmentation.

\parsection{Work Flow} Given an input image, the backbone network equipped with FPN first extracts intermediate convolutional features for downstream processing. 
Then, the object detection head predicts bounding boxes with positions as well as categories for potential instances, and prepares the cropped ROI feature for BCNet to produce segmentation masks.
The occlusion perception branch consists of the first GCN layer followed by FCN (two convolution layers), which is targeted for modeling occluding regions by jointly detecting contours and masks. 
Forming a residual connection, the distilled occlusion feature is element-wise added to the input ROI feature and passed to second GCN.
Finally, the second GCN, which has a similar structure to the first GCN, segments the occludee guided by this occlusion-aware feature and outputs contours and masks for the partially occluded instance.
\vspace{-0.2in}

\subsection{Bilayer Transformer-based Instance Segmentation}
\label{sec:bilayer_trans}
Driven by the powerful object detection paradigms of DETR~\cite{carion2020end}, transformer-based instance segmentation methods~\cite{QueryInst,dong2021solq,transfiner,cheng2021mask2former} show ever increasing performance on COCO. While these methods excels in object bounding box detection, the problem of accurately delineating each distinct object from heavy occlusions remains elusive.

We build our transformer-based BCNet based on Mask2Former~\cite{cheng2021mask2former} owing to its simple and effective architecture. In Figure~\ref{fig:framework_trans}, comparing to~\cite{cheng2021mask2former} (right part), we explicitly divide the learnable instance queries into occluder and occludee sets respectively. 
To separately model occluder and occludee information in the image, our Bilayer Transformer decoder consists of two cascaded transformer decoders, instead of using a shared one with single query group to only focus the target object (occludee). 


\parsection{Instance Queries for Occluders and Occludees} 
Transformer-based BCNet first initializes the instance queries of occludees as learnable positional embeddings. Then, to construct the occluder-occludee query pair for each image object, BCNet produces the same number of instance queries for occluders conditioned on their corresponding occludee queries. The conditional generation is based on a two-layer MLP, taking as input the query embeddings of occludees. In case of multiple occluders for an object (occludee), the occluder query group represents their grouped occlusion regions. To avoid matching conflicts, we copy bipartite matching between the occludee queries and ground truth, and then directly assign the matching correspondence to the occluder queries.

\parsection{Bilayer Transformer Decoder} Instead of solely separating input queries as occluders and occludees, comparing to conventional transformer, our Bilayer Transformer Decoder is composed of two decoders in a cascaded structure. In Figure~\ref{fig:framework_trans}, the first transformer decoder takes the instance queries of the occluders as input and predicts their object masks. Guided by occluder information from the first decoder, the second transformer decoder takes the occludee instance queries, and regresses the object masks for the target objects (occludee).
The bilayer decoder design prevents intervention between two sets of instance queries during the self-attention between input queries. Thus, the occluder query of one instance does not need to attend to the queries from the occludee set. 
However, similar to GCN-based BCNet, the overlapping information flows from the occluder decoder to occludee decoder by a residual connection.
We validate the benefit of our bilayer transformer decoder design and occlusion-aware guidance in experimental section.

\subsection{End-to-end Parameter Learning}
The whole instance segmentation framework can be trained in an end-to-end manner defined by a multi-task loss function $\mathcal{L}$ as,
\begin{equation}
\mathcal{L} = \lambda_1\mathcal{L}_{\text{Detect}} + \mathcal{L}_{\text{Occluder}} + \mathcal{L}_{\text{Occludee}},
\end{equation}
\begin{equation}
\mathcal{L}_{\text{Occluder}} = \lambda_2\mathcal{L^\prime}_{\text{Occ-B}} + \lambda_3\mathcal{L^\prime}_{\text{Occ-S}}
\label{eq:eq10}
\end{equation}
\begin{equation}
 \mathcal{L}_{\text{Occludee}} = \lambda_4\mathcal{L}_{\text{Occ-B}} + \lambda_5\mathcal{L}_{\text{Occ-S}},
 \label{eq:eq11}
\end{equation}
where~$\mathcal{L}_{\text{Occ-B}}$ and $\mathcal{L}_{\text{Occ-S}}$ denote respectively the boundary detection and mask segmentation losses in the second GCN layer for the occludee, which are similar to Eq.~\ref{eq:eq4} and Eq.~\ref{eq:eq5}.~$\mathcal{L}_{\text{Detect}}$ supervises  both the position prediction and the category classification borrowed from the FCOS~\cite{tian2019fcos} detector, 
\begin{equation}
\mathcal{L}_{\text{Detect}} = \mathcal{L}_{\text{Regression}} + \mathcal{L}_{\text{Centerness}} + \mathcal{L}_{\text{Class}},
\end{equation}
and $\lambda_1$, $\lambda_2$, $\lambda_3$, $\lambda_4$ and $\lambda_5$ are hyper-parameter weights to balance the loss functions, which are tuned to be $\{1, 0.5, 0.25, 0.5, 1.0\}$ respectively on the validation set. For transformer-based BCNet, the~$\mathcal{L}_{\text{Detect}}$ is adapted to,
\begin{equation}
\mathcal{L}_{\text{Detect}} = \mathcal{L}_{\text{Box}} + \mathcal{L}_{\text{Matching}} + \mathcal{L}_{\text{Class}},
\end{equation}
where~$\mathcal{L}_{\text{Matching}}$ denotes bipartite matching loss between predicted and ground truth objects, and~$\mathcal{L}_{\text{Box}}$ is bounding box regression loss using weighted combination of L1 loss and IoU loss following~\cite{carion2020end}.

\begin{figure*}[!h]
	\centering
	\vspace{-0.1in}
	\includegraphics[width=1.0\linewidth]{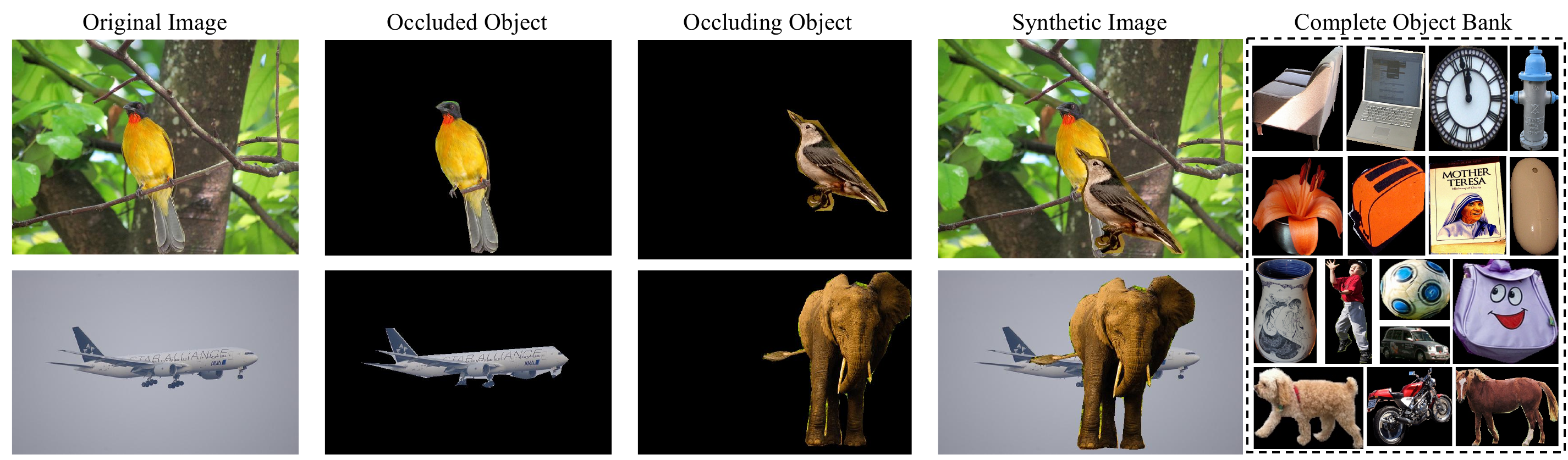}
	\vspace{-0.25in}
	\caption{Occlusion synthesis for producing \textit{Synthetic Occlusion Dataset (SOD)} by sampling both occluding and occluded instances from the collected Complete Object Bank (COB), followed by grid searching the occluded positions in the image. COB from COCO is produced by conditionally filtering out the objects with bounding boxes overlapping rate over 5\% and mask area smaller than 32$\times$32, followed by manual selection.}
	\label{fig:synthetic}
	\vspace{-0.2in}
\end{figure*}

\parsection{Training:}
During training, following Mask R-CNN~\cite{he2017mask}, GCN-based BCNet only samples RPN proposals with both highest IoU (at least larger than 0.5) to the GT boxes and high classification confidence for mask head training. On COCO, for each sampled proposal box, its occludee is simplified as the target object belonging to its best matched GT box.
For training the first GCN layer of BCNet, since partial occlusion cases only occupy a small fraction compared to the complete objects in COCO, we filter out part of the non-occluded ROI proposals to keep occlusion cases taking up 50\% for balance sampling. SGD with momentum is employed for training 90K iterations which starts with 1K constant warm-up iterations. The batch size is set to 16 and initial learning rate is 0.01.  In ablation study, ResNet-50-FPN~\cite{he2016deep} is used as backbone and the input images are resized without changing the aspect ratio by keeping the shorter side and longer side of no more than 600 and 900 pixels respectively. For leaderboard comparison, we adopt the scale-jitter where the shorter image side is randomly sampled from [640, 800] following 3$\times$ schedule in~\cite{lee2019centermask,chen2019tensormask,bolya2019yolact}. 
For the transformer-based BCNet, we follow the same training schedules and setting in~\cite{cheng2021mask2former}, where we train the model for 50 epochs with a batch size of 16 and large-scale jittering~\cite{ghiasi2021simple}. For fair comparison, transformer-based BCNet follows the same segmentation loss in Mask2Former \textit{without} boundary detection mentioned in Eq.~\ref{eq:eq10} and Eq.~\ref{eq:eq11}, increasing the training time of Mask2Former by 20\%. Since there are no RoI proposals in Mask2Former, we adopt the complete GT mask annotation to determine the occluder pixels, i.e., the union/grouping of objects spatially neighboring to the target occludee. We take 100 instance queries per image for occluders and occludees respectively.

\parsection{Inference:} During inference, the mask head in GCN-based BCNet predicts masks for the occluded target object in the high-score box proposals (no more than 50) generated by the FCOS detector, where the first GCN layer only produces occlusion-aware feature as input for the second GCN. 

\vspace{-0.1in}

\section{Synthetic Occlusion Dataset}
In this section, we provide details about the proposed Synthetic Occlusion Dataset (SOD) for instance
segmentation. SOD facilitates occluded objects understanding.


\parsection{Occlusion Synthesis Process}
As shown in Figure~\ref{fig:synthetic}, to diversify the occlusion patterns, we construct the large-scale Synthetic Occlusion Dataset (SOD) by sampling both occluding and occluded instances from the Complete Object Bank (COB) following uniform class distribution. COB consists of images for \textit{non-occluded single} object with corresponding complete mask and contour annotation, which has 80 categories with total instances number over 60,000. Then, a synthetic image based on the original image corresponding to the occluded target is produced by placing the occluding instance at a random image position (generated by grid search) which satisfies the object overlapping rate between 0.2 to 0.5. The synthetic occlusion dataset contains 100K such occluded images with \textbf{amodal} contours/masks for {\em both} occluding and partially occluded objects. We show the benefit of additionally training BCNet on SOD in Table~\ref{tab:occ_split}.

\section{Experiments}\label{sec:experiment}

\subsection{Experimental Setup}

\parsection{COCO and COCO-OCC}
We conduct experiments on COCO dataset~\cite{lin2014microsoft}, where we train on 2017{\it train} (115k images) and evaluate results on both 2017{\it val} and 2017{\it test-dev} using the standard metrics. For further investigating segmentation performance with occlusion handling, we propose a subset split, called COCO-OCC, which contains 1,005 images extracted from the validation set (5k images) where the overlapping ratio between the bounding boxes of objects is at least 0.2. Segmenting COCO-OCC with highly overlapping objects is much more difficult than 2017{\it val}, where we observe a performance gap around 3.0$AP$ for the same model in the experiment section. Besides, we also validate the synthetic SOD dataset on COCO-OCC.

\parsection{KINS and COCOA}
We also evaluate BCNet on two amodal instance segmentation benchmarks: 
(1) \textbf{KINS}~\cite{qi2019amodal}, built on the original KITTI~\cite{geiger2012we}, is the largest amodal segmentation benchmark for traffic scenes with both annotated amodal and modal masks for instances. BCNet is trained on the training split (7,474 images and 95,311 instances) and tested on the testing split (7,517 images and 92,492 instances) following the setting in~\cite{qi2019amodal}.
(2) \textbf{COCOA}~\cite{zhu2017semantic} is a subpart of COCO~\cite{lin2014microsoft}, where we train BCNet on the official training split (2,500 images) and test on the validation split (1,323 images). Note that each instance has no class label and we only use the modal and amodal mask labels for COCOA.

\parsection{Youtube-VIS, OVIS and BDD100K MOTS}
 We further evaluate the GCN-based BCNet on three large VIS/MOTS benchmarks: 1) \textbf{YTVIS}~\cite{yang2019video} is a Video Instance Segmentation (VIS) benchmark, which contains 2,883 videos with 131k annotated object instances of 40 categories. We also report the results of BCNet on OVIS~\cite{qi2021occluded}, a new VIS dataset on occlusion learning; 2) \textbf{OVIS} has 607, 140 and 154 videos for training, validation and test respectively. To evaluate BCNet in video instance segmentation, we only replace the frame-level mask head of Mask Track R-CNN~\cite{yang2019video} and CMTrack RCNN~\cite{qi2021occluded} while leaving the other model components unchanged; 3) \textbf{BDD100K MOTS}~\cite{bdd100k} is a large-scale Multiple Object Tracking and Segmentation (MOTS) dataset of BDD100K~\cite{bdd100k}, which includes 154 videos (30,817 images) for training, 32 videos (6,475 images) for validation, and 37 videos (7,484 images) for testing. We integrate the mask head of BCNet into PCAN~\cite{pcan} and adopt the well-established MOTS metrics~\cite{voigtlaender2019mots} for results comparison. BDD100K covers the self-driving scenario while YTVIS and OVIS have more diverse object categories.

\subsection{Ablation Study}

\parsection{Effect of Explicit Occlusion Modeling} We validate the efficacy of different components proposed for explicit occlusion modeling on the first GCN layer. Table~\ref{tab:fist_GCN} tabulates the  quantitative comparison: 1) Baseline: BCNet with no explicit occlusion modeling targets; 2) modeling segmentation masks for occluding regions (\textbf{occluder}); 3)  modeling contours of the occluding regions; 4) \textbf{joint} occlusion modeling on both masks and contours.   Compared to the baseline, joint occlusion modeling produces the most obvious improvement especially for the heavy occlusion cases, which promotes mask $AP$ on the standard validation set from 32.65 to 33.43, and the $AP$ on the proposed COCO-OCC split is increased from 29.04 to 30.37.

\begin{table}[!h]
	\caption{Effect of the first GCN for  occlusion modeling by predicting contours and masks on COCO with ResNet-50-FPN model.}
	\centering
	\resizebox{0.85\linewidth}{!}{
		\begin{tabular}{c | c | c | c | c | c }
			\toprule
			\multicolumn{2}{c|}{Occlusion (Occluder) Modeling} & \multicolumn{2}{c|}{COCO-OCC} & \multicolumn{2}{c}{COCO} \\
			\cline{1-6} 
			Contour & Mask & $AP$ & $AP_{50}$ & $AP$ & $AP_{50}$\\
			\midrule
			&  & 29.04 & 49.22 & 32.65 & 52.39 \\
			& \checkmark & 29.65 & 49.42 & 33.25 & 52.82 \\
			\checkmark  & & 30.18 & 49.94 & 33.41 & 53.02 \\
			\checkmark  & \checkmark  & \textbf{30.37} & \textbf{50.40} & \textbf{33.43} &  \textbf{53.12} \\
			\bottomrule
		\end{tabular}
	}
	\vspace{-0.1in}
	\label{tab:fist_GCN}
\end{table}

\parsection{Effect of Bilayer Occluder-occludee Modeling} Built on the first GCN layer with explicit occlusion modeling, we further validate the second GCN layer in Table~\ref{tab:second_GCN}, which demonstrates the importance of~\textit{occlusion-aware} feature \textit{guidance} for the second GCN layer to segment target object (\textbf{occludee}) by boosting 1.23 $AP$ on COCO-OCC, and 1.06 $AP$ on COCO respectively. Table~\ref{tab:bilayer} shows the results comparison on adopting the proposed~\textit{bilayer structure} and existing direct regression model with single layer. On the COCO-OCC split, bilayer GCN improves $AP$ from 29.63 to 30.68 compared to single GCN, and bilayer FCN boosts the performance of single FCN from 28.43 to 30.12.

\begin{table}[!h]
	\caption{Effect of the second GCN for detecting occludee contours for final mask prediction \textbf{\textit{guided}} by the output of first GCN.}
	\centering
	\resizebox{0.95\linewidth}{!}{
		\begin{tabular}{c | c | c | c | c | c | c }
			\toprule
			\multicolumn{3}{c|}{Target (Occludee) Modeling} & \multicolumn{2}{c|}{COCO-OCC} & \multicolumn{2}{c}{COCO} \\
			\cline{1-7} 
			Guidance & Contour & Mask & $AP$ & $AP_{50}$ & $AP$ & $AP_{50}$\\
			\midrule
			& & \checkmark & 29.45 & 49.73 & 32.56 & 52.21 \\
			\checkmark & & \checkmark & 30.37 & 50.40 & 33.43 & 53.12 \\
			\checkmark & \checkmark  & \checkmark  & \textbf{30.68} & \textbf{50.62} & \textbf{33.62} & \textbf{53.26} \\
			\bottomrule
		\end{tabular}
	}
	\vspace{-0.1in}
	\label{tab:second_GCN}
\end{table}

\parsection{Using FCN or GCN?} Table~\ref{tab:bilayer} also reveals the advantage of GCN over FCN, where GCN achieves consistent superior performance both in the singe layer and bilayer structure.
We also compute parameters number of each model and find that although GCN has more trainable parameters, the increased model size is acceptable compared to performance gain, because the feature size of input ROI has been down-sampled to only 14$\times$14 (spatial size) with 256 channels.

\begin{table}[!h]
	\caption{Effect of~\textbf{bilayer structure} using \textbf{GCN vs. FCN} implementation.}
	\centering
	\resizebox{0.95\linewidth}{!}{
		\begin{tabular}{c | c | c | c | c | c | c | c}
			\toprule
			\multirow{2}{*}{Structure} & \multirow{2}{*}{FCN} & \multirow{2}{*}{GCN} & \multicolumn{2}{c|}{COCO-OCC} & \multicolumn{2}{c|}{COCO} & \multirow{2}{*}{Params}\\
			\cline{4-7} 
			& & & $AP$ & $AP_{50}$ & $AP$ & $AP_{50}$ &\\
			\midrule
			\multirow{2}{*}{Single Layer} & \checkmark  & & 28.43 & 48.24 & 33.01 & 52.62 & 51.0M\\
			& & \checkmark & 29.63 & 49.59 & 33.14 & 52.81 & 51.4M\\
			\midrule
			\multirow{2}{*}{Bilayer} & \checkmark  & & 30.12 & 49.04 & 33.16 & 52.80 & 53.4M\\
			& & \checkmark & \textbf{30.68} & \textbf{50.62} & \textbf{33.62} & \textbf{53.26} & 54.0M\\
			\bottomrule
		\end{tabular}
	}
	\vspace{-0.1in}
	\label{tab:bilayer}
\end{table}

\begin{table}[!h]
	\caption{Influence of the object detector (FCOS vs. Faster R-CNN vs. Query-based detector~\cite{cheng2021mask2former}) on BCNet.}
	\centering
	\resizebox{1.0\linewidth}{!}{
		\begin{tabular}{ l | c | c | c | c | c}
			\toprule
			\multirow{2}{*}{Model} & \multicolumn{2}{c|}{COCO-OCC} & \multicolumn{2}{c|}{COCO} & \multirow{2}{*}{Params} \\
			\cline{2-5} 
			& $AP$ & $AP_{50}$ & $AP$ & $AP_{50}$ \\
			\midrule
			FCOS~\cite{lee2019centermask} + Baseline  & 28.43 & 48.24 & 33.01 & 52.62 & 51.0M\\
			FCOS~\cite{tian2019fcos} + Ours & \textbf{30.68} & \textbf{50.62} & \textbf{33.62} & \textbf{53.26} & 54.0M\\
			\midrule
			Faster R-CNN~\cite{he2017mask} + Baseline & 29.67 & 49.95 & 33.45 & 53.70 & 60.0M\\
			Faster R-CNN~\cite{ren2015faster} + Ours & \textbf{31.71} & \textbf{51.15} & \textbf{34.61} & \textbf{54.41} & 63.2M \\
			\midrule
			Query-based Detector~\cite{cheng2021mask2former} + Baseline & 39.23 & 50.62 & 41.13 & 62.50 & 81.6M \\
			Query-based Detector~\cite{cheng2021mask2former} + Ours & \textbf{41.67} & \textbf{52.03} & \textbf{42.51} & \textbf{64.23} & 89.7M \\
			\bottomrule
		\end{tabular}
	}
	\vspace{-0.05in}
	\label{tab:detector}
\end{table}

\begin{figure}[!h]
	\centering
	\includegraphics[width=1.0\linewidth]{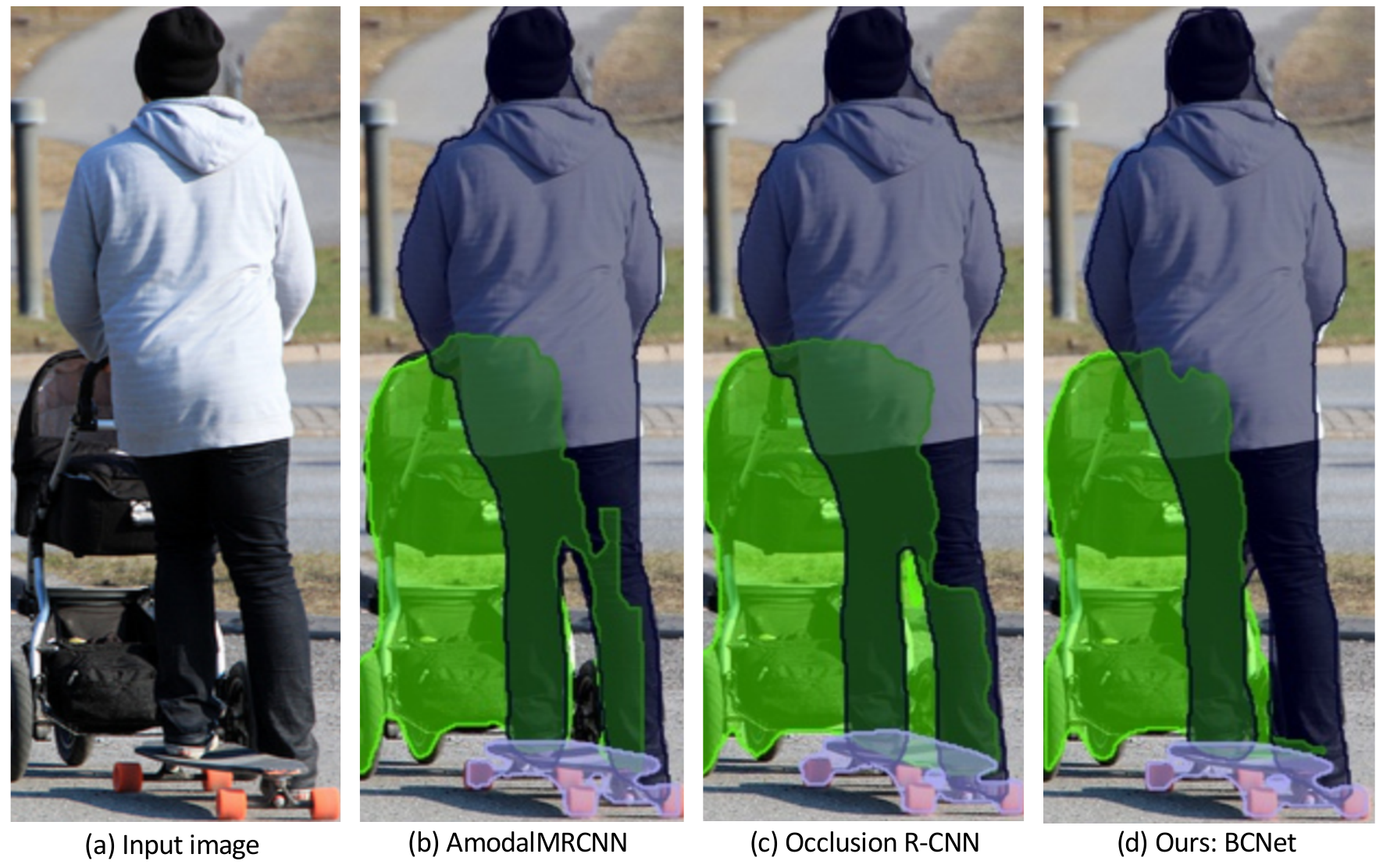}
	\vspace{-0.25in}
	\caption{Qualitative results comparison of the \textbf{amodal }mask predictions on \textbf{COCOA}~\cite{zhu2017semantic} by AmodalMRCNN~\cite{follmann2019learning}, ORCNN~\cite{follmann2019learning} and our method using ResNet-50, where BCNet hallucinates a more reasonable shape for the baby carriage without producing a large portion of segmentation error. We remove the ``stuff'' background for more clarity. }
	\label{fig:amodal_example1}
	\vspace{-0.1in}
\end{figure}

\begin{table*}[!t]
        \vspace{-0.1in}
	\caption{Effect of the~\textbf{Bilayer Transformer Decoder} for the transformer-based BCNet.}
	\vspace{-0.1in}
	\centering
	\resizebox{0.95\linewidth}{!}{
		\begin{tabular}{c | c | c | c | c | c | c | c | c | c |c}
			\toprule
			\multicolumn{4}{c|}{Transformer-based BCNet} & \multicolumn{2}{c|}{COCO-OCC} & \multicolumn{2}{c|}{COCO} & \#params. & FLOPs & fps\\
			Shared decoder (100Q) & Shared decoder (200Q) & Bi-decoder (200Q) & Occlusion-guidance & $AP$ & $AP_{50}$ & $AP$ & $AP_{50}$ & & &\\
			\midrule
			\checkmark & & & & 38.67 & 58.73 & 41.51 & 61.73 & 44.0M & 226G & 8.6\\
			 & \checkmark & & & 39.01 & 59.90 & 41.82 & 62.14 & 44.0M & 356G & 8.2\\
			 &  & \checkmark & & 40.17 & 61.20 & 42.62 & 63.02 & 53.8M & 361G & 8.0\\
			 &  & \checkmark  & \checkmark & \textbf{41.23} & \textbf{62.12} & \textbf{43.21} & \textbf{64.21} & 53.8M & 362G & 8.0\\
			\bottomrule
		\end{tabular}
	}
	\vspace{-0.15in}
	\label{tab:bilayer_decoder}
\end{table*}

\begin{table*}[!h]
	\begin{minipage}[t]{0.33\linewidth}
		\caption{Results on the COCOA dataset.}
		\centering
		\resizebox{1.0\linewidth}{!}{
			\begin{tabular}{l | c | c | c}
				\toprule
				Model & $AP_{all}$ & $AP_{t}$ & $AP_{s}$\\
				\midrule
				AmodalMask~\cite{zhu2017semantic} & 5.7 & 5.9 & 0.8 \\
				AmodalMRCNN~\cite{follmann2019learning} & 21.51 & 21.09 & 9.0 \\
				ORCNN~\cite{follmann2019learning} & 20.32 & 20.63 & 7.8 \\
				\midrule
				\textbf{BCNet} & \textbf{23.09} & \textbf{22.72} & \textbf{9.53} \\
				\bottomrule
			\end{tabular}
		}
		\label{tab:cocoa}
	\end{minipage}
	\hfill
	\begin{minipage}[t]{0.33\linewidth}
		\caption{Results on the KINS dataset.}
		\centering
		\resizebox{0.95\linewidth}{!}{
			\begin{tabular}{l | c | c }
				\toprule
				Model & $AP_{Det}$ & $AP_{Seg}$ \\
				\midrule
				Mask R-CNN~\cite{follmann2019learning} & 26.97 & 24.93 \\
				Mask R-CNN + ASN~\cite{qi2019amodal} & 27.86 & 25.62 \\
				PANet~\cite{liu2018path} & 27.39 & 25.99 \\
				PANet + ASN~\cite{qi2019amodal} & 28.41 & 26.81 \\
				\midrule
				\textbf{BCNet} & \textbf{28.87} & \textbf{27.30} \\
				\bottomrule
			\end{tabular}
		}
		\label{tab:kins}
	\end{minipage}
	\hfill
	\begin{minipage}[t]{0.33\linewidth}
		\caption{Results on COCO-OCC split.}
		\centering
		\resizebox{0.8\linewidth}{!}{
			\begin{tabular}{l | c | c }
				\toprule
				Model & $AP$ & $AP_{50}$ \\
				\midrule
				Mask R-CNN~\cite{he2016deep} & 29.67 & 49.95 \\
				CenterMask~\cite{lee2019centermask} & 29.05 & 49.07 \\
				MS R-CNN~\cite{huang2019mask} & 30.32 & 50.01 \\
				\midrule
				\textbf{Ours} & 31.71 & 51.15\\
				\textbf{Ours + SOD} & \textbf{32.89} & \textbf{53.25}\\
				\bottomrule
			\end{tabular}
		}
		\label{tab:occ_split}
	\end{minipage}
	\vspace{-0.1in}
\end{table*}

\begin{table}[!t]
        \vspace{-0.3in}
	\centering%
		\caption{Results on the OCHuman~\cite{zhang2019pose2seg} \textit{val} using R50-FPN.}\vspace{-1mm}
	\resizebox{0.3\textwidth}{!}{%
	\begin{tabular}{lcccccc}
					\toprule
					Method & AP & AP$_{M}$ & AR$_{L}$ \\
					\midrule
					Mask R-CNN~\cite{he2017mask} & 16.3 & 19.4 & 11.3 \\
					\textbf{BCNet} & \textbf{20.6} & \textbf{23.3} & \textbf{13.8}  \\ 
					\bottomrule
			\end{tabular}}
		\label{tab:vis}
		\vspace{-5mm}
	\end{table}
 
\parsection{Effect of Bilayer Transformer Decoder}
Table~\ref{tab:bilayer_decoder} tabulates the effect of our transformer-based BCNet with Bilayer Transformer Decoder. Compared to the standard shared transformer decoder~\cite{cheng2021mask2former} with single set of instance queries (200), our bilayer transformer decoder training for 36 epochs with both occluder and occludee queries respectively improves 1.50 $AP$ on COCO-OCC, and 1.01 $AP$ on COCO. By further injecting the occlusion-aware guidance from the first transformer decoder to the second decoder, the mask AP can respectively be boosted from 40.17 to 41.23 on COCO-OCC, and from 42.62 to 43.21 on COCO validation set.

\parsection{Influence of Object Detector} 
To investigate the influence of object detectors to BCNet, besides using one-stage detector FCOS~\cite{tian2019fcos}, we also use representative two-stage and query-based detectors Faster R-CNN~\cite{ren2015faster} to perform experiments.  As shown in Table~\ref{tab:detector}, the performance gain brought by BCNet is consistent, with an improvement of 2.23 (for FCOS), 2.04 (for Faster R-CNN) mask $AP$ on COCO-OCC respectively. The query-based BCNet improves 1.38 mask AP on COCO, and 2.44 mask AP on COCO-OCC. Note the baseline in one/two-stage detector denotes mask head design in Mask R-CNN, while the baseline in query-based detector denotes the mask head design of Mask2Former.

\subsection{Performance Comparison and Analysis}

\begin{figure*}[!t]
	\centering
	\includegraphics[width=1.0\linewidth]{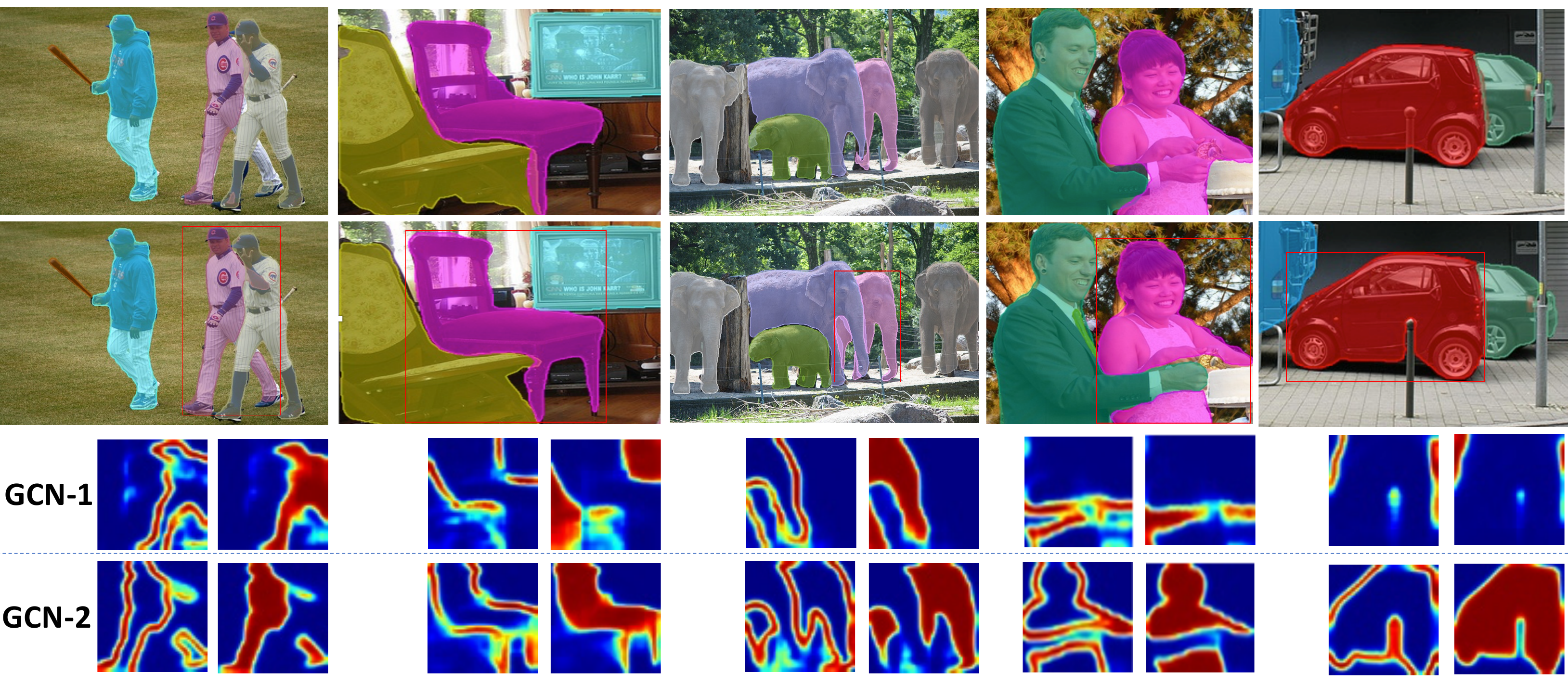}
	\vspace{-0.3in}
	\caption{Qualitative instance segmentation results of CenterMask~\cite{lee2019centermask} (top row) and our BCNet (middle row) on \textbf{COCO}~\cite{lin2014microsoft}, both using ResNet-101-FPN and \textbf{FCOS detector}~\cite{tian2019fcos}. The bottom row visualizes squared heatmap of contour and mask predictions by the two GCN layers for the occluder and occludee in the same~\textbf{ROI region} specified by the red bounding box, which also makes the final segmentation result of BCNet more explainable 
	than previous methods.}
	\label{fig:qualitative}
	\vspace{-0.1in}
\end{figure*}

\begin{figure*}[!t]
	\centering
	\includegraphics[width=1.0\linewidth]{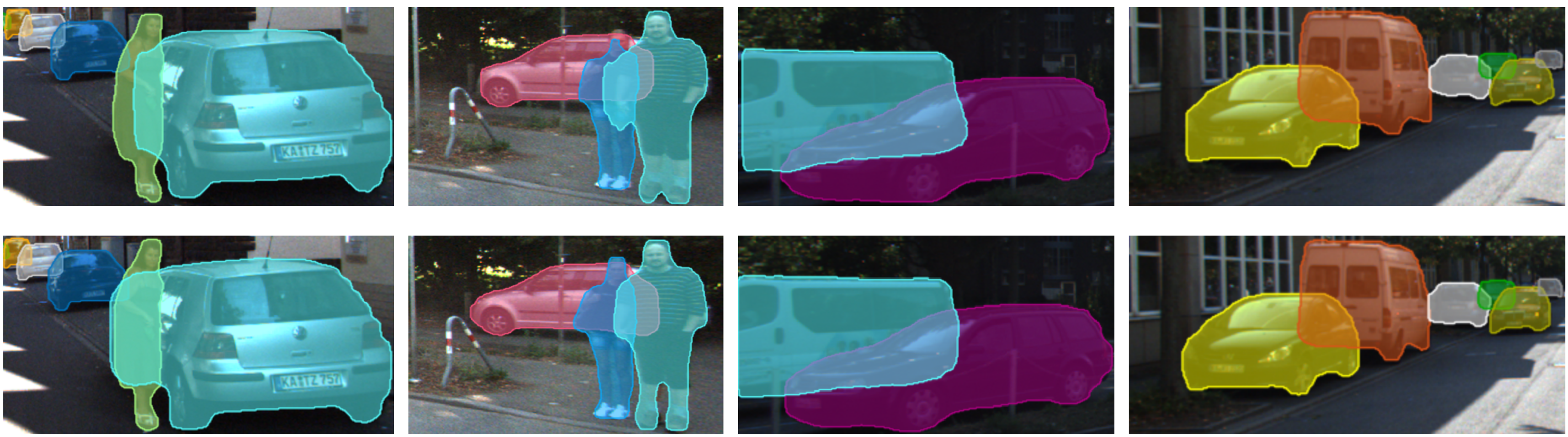}
	\vspace{-0.2in}
	\caption{Qualitative \textbf{amodal} results comparison between Mask R-CNN + ASN module~\cite{qi2019amodal} (top row) and our BCNet (bottom row) for the mask predictions on~\textbf{KINS} test set~\cite{qi2019amodal}, both using ResNet-101-FPN  and Faster R-CNN detector~\cite{ren2015faster}, where the mask shape of the \textbf{invisible/occluded regions} are more reasonably estimated by BCNet.}
	\label{fig:kins_compare}
	\vspace{-0.1in}
\end{figure*}

\begin{table*}[!t]
	\caption{Comparison with SOTA methods on COCO {\it test-dev} set. Mask AP is reported and all entries are single-model results. Note that HTC~\cite{chen2019hybrid} adopts 3-stage cascade refinement with multiple object detectors and mask heads. All of the methods are trained on COCO \emph{train2017}.}
	\vspace{-0.25in}
	\begin{center}{\small
			\resizebox{0.9\linewidth}{!}{
				\begin{tabular}{ccc|c|ccc|ccc}
					\hline
					& Method & Backbone & Type & $AP$ & $AP_{50}$ & $AP_{75}$ & $AP_{S}$ & $AP_{M}$ & $AP_{L}$ \\ 
					\hline
					\multirow{4}{*}{} & Mask R-CNN~\cite{he2017mask} & R50-FPN & Two-stage & 35.6 & 57.6 & 38.1 & 18.7 & 38.3 & 46.6 \\ 
					& PANet~\cite{liu2018path} & R50-FPN & Two-stage & 36.6 & 58.0 & 39.3 & 16.3 & 38.1 & \textbf{52.4} \\
					& \textbf{BCNet + Faster R-CNN~\cite{ren2015faster}} & R50-FPN & Two-stage & \textbf{38.4} & \textbf{59.6} & \textbf{41.5} & \textbf{21.9} & \textbf{40.9} & 49.3 \\
					\hline
					& Mask R-CNN~\cite{he2017mask} & R101-FPN & Two-stage & 37.0 & 59.2 & 39.5 & 17.1 & 39.3 & 52.9 \\ 
					& MaskLab~\cite{chen2018masklab} & R101-FPN & Two-stage & 37.3 & 59.8 & 39.6 & 19.1 & 40.5 & 50.6 \\
					& Mask Scoring R-CNN~\cite{huang2019mask} & R101-FPN & Two-stage & 38.3 & 58.8 & 41.5 & 17.8 & 40.4 & 54.4 \\
					& BMask R-CNN~\cite{ChengWHL20} & R101-FPN & Two-stage & 37.7 & 59.3 & 40.6 & 16.8 & 39.9 & \textbf{54.6} \\
					& HTC~\cite{chen2019hybrid} & R101-FPN & Two-stage & 39.7 & \textbf{61.8} & 43.1 & 21.0 & 42.2 & 53.5 \\
					& \textbf{BCNet + Faster R-CNN~\cite{ren2015faster}}      & R101-FPN & Two-stage & {\bf 39.8}    & 61.5 & {\bf 43.1} & {\bf 22.7} & {\bf 42.4} & 51.1  \\
					\hline
					\hline
					\multirow{5}{*}{}& YOLACT~\cite{bolya2019yolact}       & R101-FPN & One-stage & 31.2 & 50.6 & 32.8 & 12.1 & 33.3 & 47.1 \\
					& TensorMask~\cite{chen2019tensormask}   & R101-FPN & One-stage & 37.1 & 59.3 & 39.4 & 17.4 & 39.1 & 51.6 \\
					& ShapeMask~\cite{kuo2019shapemask}      & R101-FPN & One-stage & 37.4 & 58.1 & 40.0 & 16.1 & 40.1 & 53.8 \\ 
					& CenterMask~\cite{lee2019centermask}    & R101-FPN & One-stage & 38.3 & - & - & 17.7 & 40.8 & {\bf 54.5} \\
					& BlendMask~\cite{chen2020blendmask}    & R101-FPN & One-stage & 38.4 & 60.7 & 41.3 & 18.2 & 41.5 & 53.3 \\
					& \textbf{BCNet + FCOS~\cite{tian2019fcos}}  & R101-FPN & One-stage & {\bf 39.6}    & {\bf 61.2} & {\bf 42.7} & {\bf 22.3} & {\bf 42.3} & 51.0  \\
					\hline
					\hline
					\multirow{5}{*}{}& 
					ISTR~\cite{hu2021ISTR}  & R50-FPN & Query-based & 38.6 & - & - & 22.1 & 40.4 & 50.6 \\
					& QueryInst~\cite{QueryInst} & R50-FPN & Query-based & 39.9 & 62.2 & 43.0 & 22.9 & 41.7 & 51.9  \\
					& SOLQ~\cite{dong2021solq} & R50-FPN & Query-based & 39.7 & - & - & 21.5 & 42.5 & 53.1 \\ 
					& Mask Transfiner~\cite{transfiner} & R50-FPN & Query-based & 41.6 & 63.9 & 45.5 & 24.2 & 44.6 & 55.2 \\
					& Mask2Former~\cite{cheng2021mask2former} & R50-FPN & Query-based & 43.6 & 66.5 & 47.9 & 23.5 & 47.4 & 64.1 \\
					& \textbf{Transformer-based BCNet} & R50-FPN  & Query-based & \textbf{44.6} & \textbf{68.1} & \textbf{48.7} & \textbf{24.1} & \textbf{47.7} & \textbf{66.7}  \\
					\hline
		\end{tabular}}}
	\end{center}
	\label{table:fully}
	\vspace{-0.2in}
\end{table*}

\begin{table}[!t]
	\centering%
		\caption{State-of-the-art comparison of BCNet built on Mask Track R-CNN~\cite{yang2019video} on the YouTube-VIS validation set, using ResNet-50 as backbone. Results are reported in terms of mask accuracy (AP) and recall (AR).}
	\resizebox{0.5\textwidth}{!}{%
	\begin{tabular}{lcccccc}
					\toprule
					Method & AP & AP$_{50}$ & AP$_{75}$ & AR$_{1}$ & AR$_{10}$ \\
					\midrule
					OSMN~\cite{yang2018efficient} & 23.4 & 36.5 & 25.7 & 28.9 & 31.1 \\
					FEELVOS~\cite{voigtlaender2019feelvos} & 26.9 & 42.0 & 29.7 & 29.9 & 33.4 \\
					DeepSORT~\cite{wojke2017simple} & 26.1 & 42.9 & 26.1 & 27.8 & 31.3 \\
					\midrule
					MaskTrack R-CNN~\cite{yang2019video} & 30.3 & 51.1 & 32.6 & 31.0 & 35.5 \\
					MaskTrack R-CNN~\cite{yang2019video} \textbf{+ BCNet} & \textbf{32.4} & \textbf{53.9} & \textbf{34.0} & \textbf{33.9} & \textbf{39.1} \\ 
					\bottomrule
			\end{tabular}}
		\label{tab:vis}
		\vspace{-2mm}
	\end{table}
	
	\begin{table}[t]
	\centering%
		\caption{State-of-the-art comparison of BCNet built on CMTrack RCNN~\cite{qi2021occluded} on the OVIS validation set, using ResNet-50 as backbone. Results are reported in terms of mask accuracy (AP) and recall (AR).}
	\resizebox{0.5\textwidth}{!}{%
	\begin{tabular}{lccccc}
					\toprule
					Method & AP & AP$_{50}$ & AP$_{75}$ & AR$_{1}$ & AR$_{10}$ \\
					\hline
					MaskTrack~\cite{yang2019video} & 10.8 & 25.3 & 8.5 & 7.9 & 14.9 \\
					SipMask~\cite{CaoSipMask_ECCV_2020} & 10.2 & 24.7 & 7.8 & 7.9 & 15.8 \\
					QueryInst~\cite{QueryInst} & 14.7 & 34.7 & 11.6 & 9.0 & 21.2 \\ 
					CrossVIS~\cite{Yang_2021_ICCV} & 14.9 & 32.7 & 12.1 & 10.3 & 19.8 \\
					STMask~\cite{STMask-CVPR2021} & 15.4 & 33.8 & 12.5 & 8.9 & 21.3 \\ 
					\midrule
					CMTrack RCNN~\cite{qi2021occluded} & 15.4 & 33.9  & 13.1 & 9.3 & 20.0 \\ 
					CMTrack RCNN~\cite{qi2021occluded} \textbf{+ BCNet} & \textbf{17.1} & \textbf{35.8} & \textbf{14.2} & \textbf{10.9} & \textbf{21.3} \\
					\bottomrule
			\end{tabular}}
		\label{tab:ovis}
		\vspace{-4mm}
	\end{table}

	
	\begin{figure*}[!t]
	\centering
	\includegraphics[width=1.0\linewidth]{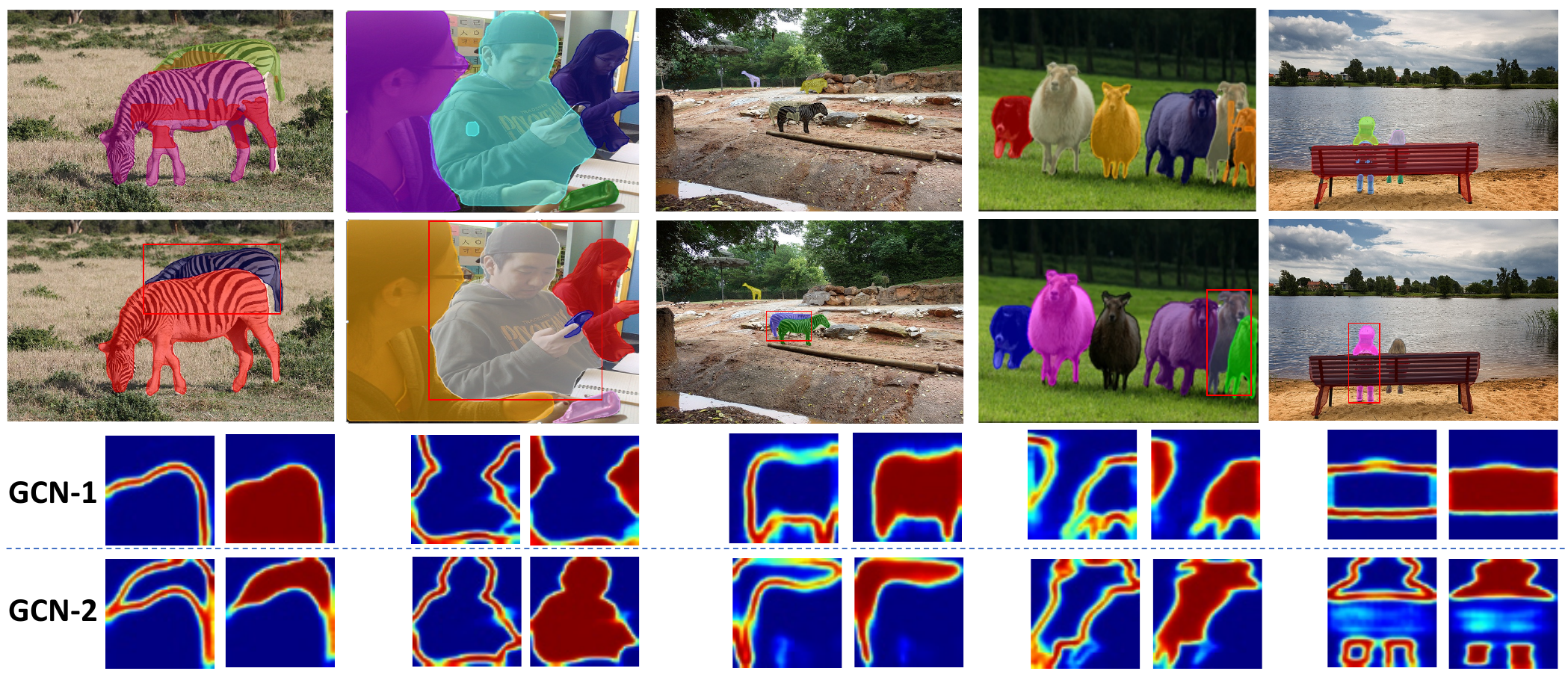}
	\vspace{-0.3in}
	\caption{Qualitative results of Mask Scoring R-CNN~\cite{huang2019mask} (top row) and our BCNet (middle row) on \textbf{COCO} {\it test-dev} set, both using ResNet-101-FPN and \textbf{Faster R-CNN~\cite{ren2015faster}}. The bottom row visualizes squared heatmap of contour and mask predictions by the two GCN layers for the occluder and occludee in the same~\textbf{ROI region} specified by the  red bounding box, which also makes the final segmentation result of BCNet more explainable than previous methods. }
	\label{fig:qualitative_score}
	\vspace{-0.2in}
\end{figure*}

\begin{figure*}[!t]
	\centering
	\includegraphics[width=1.0\linewidth]{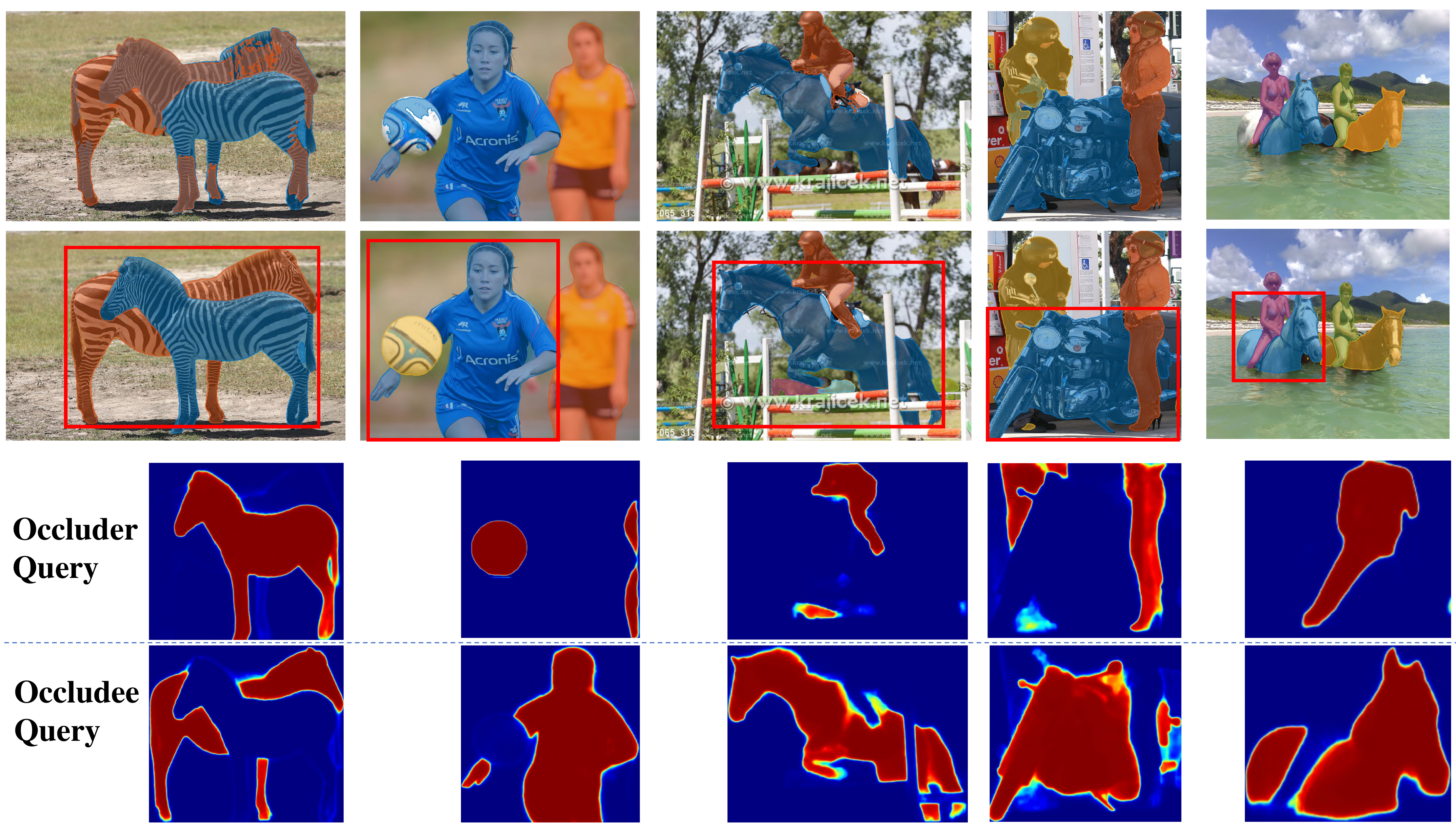}
	\vspace{-0.3in}
	\caption{Qualitative instance segmentation results of \textbf{transformer-based} BCNet with single transformer decoder (top row) and bilayer transformer structure (middle row) on \textbf{COCO}~\cite{lin2014microsoft}, both using ResNet-50-FPN. The bottom row visualizes squared heatmap of mask predictions by the occluder and occludee queries in the same region specified by the {\color{red} red} bounding box.}
	\label{fig:qualitative_transformer}
	\vspace{-0.1in}
\end{figure*}


\parsection{Comparison with Amodal Segmentation Methods}
Table~\ref{tab:cocoa} and Table~\ref{tab:kins} compare BCNet with other SOTA amodal segmentation methods on both the COCOA~\cite{zhu2017semantic} and KINS~\cite{qi2019amodal} datasets, where: 1) AmodalMask~\cite{zhu2017semantic} directly predicts amodal masks from image patches; 2) Occlusion RCNN (ORCNN)~\cite{follmann2019learning} is an extension of Mask R-CNN with both amodal and modal mask heads; 3) ASN module~\cite{qi2019amodal} contains additional occlusion classification branch and multi-level coding. 
Compared to these occlusion handling approaches, our bilayer GCN with cascaded structure still performs favorably against the state-of-the-art methods, which shows the effectiveness of BCNet in decoupling overlapping objects and mask completion under the amodal segmentation setting.
Figure~\ref{fig:amodal_example1} and Figure~\ref{fig:kins_compare} show the qualitative comparison on COCOA and KINS respectively.

\parsection{Evaluation on Occluded Images}
We adopt COCO-OCC split to compare the occlusion handling ability of BCNet with other methods on images with highly overlapping objects. As shown in Table~\ref{tab:occ_split}, our BCNet with Faster R-CNN detector has 31.71 \textit{AP} \textit{vs.} 30.32 for the Mask Scoring R-CNN~\cite{huang2019mask}. By further training BCNet on the synthetic occlusion dataset (SOD), the performance of~\textit{AP} and \textit{AP}$_{50}$ is significantly promoted to 32.89 and 53.25 respectively, which shows the advantage brought by this new dataset.
We also evaluate GCN-based BCNet on OCHuman~\cite{zhang2019pose2seg}. The mask AP for Mask R-CNN (baseline) is 16.3. Although not specifically designed for handling human occlusions, our BCNet reaches 20.6 mask AP without any keypoint/pose usage, achieving large 4.3 mask AP improvement.

\parsection{Comparison with SOTA Methods}
Table~\ref{table:fully} compares BCNet with  state-of-the-art instance segmentation methods on COCO dataset.
BCNet achieves consistent improvement on different backbones and object detectors, demonstrating its effectiveness by outperforming both PANet~\cite{liu2018path} and Mask Scoring R-CNN~\cite{huang2019mask} by 1.5 mask \textit{AP} using two-stage detector Faster R-CNN, exceeding CenterMask~\cite{lee2019centermask} by 1.3~\textit{AP} using one-stage detector FCOS, improving Mask2Former~\cite{cheng2021mask2former} by 0.9~\textit{AP} using \textbf{query-based} detector. Our single two-stage based model achieves comparable result with HTC~\cite{chen2019hybrid}, which uses a 3-stage cascade refinement with multiple object detectors and mask heads, and far more parameters. Without bells and whistles, our transformer-based BCNet achieves 44.6 mask AP only using R50-FPN as backbone.

\begin{figure*}[!t]
	\centering
	\includegraphics[width=1.0\linewidth]{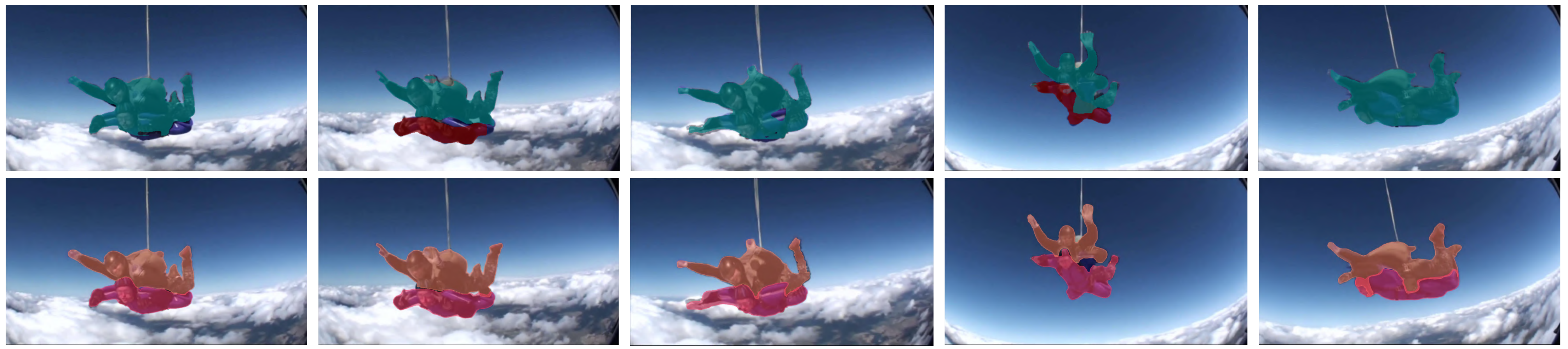}
	\vspace{-0.25in}
	\caption{Qualitative results comparison between Mask Track R-CNN~\cite{yang2019video} (top row) and Mask Track R-CNN~\cite{yang2019video} + BCNet (bottom row) using R50-FPN as backbone on YTVIS validation set. BCNet produces more accurate segmentation results inside the overlapping regions between the two tandem skydivers, by replacing the frame-level mask head of Mask Track R-CNN.} 
	\label{fig:vis}
	\vspace{-0.2in}
\end{figure*}

\begin{table}[t]
	\centering%
		\caption{State-of-the-art comparison of BCNet built on PCAN~\cite{pcan} on the BDD100K segmentation tracking validation set. I: ImageNet. C: COCO. S: Cityscapes. B: BDD100K. "-fix" means adopting the pretrained model from the BDD100K tracking set, fixing the existing parts, and only training the added mask head.}\vspace{-2mm}
	\resizebox{0.5\textwidth}{!}{%
				\begin{tabular}{lcccccc}
					\toprule
					Method & Online & mMOTSA$\uparrow$ & mMOTSP$\uparrow$ & mIDF$\uparrow$ & ID sw.$\downarrow$  & mAP$\uparrow$ \\
					\midrule
					SortIoU  & \checkmark & 10.3 & 59.9 & 21.8 & 15951 & 22.2 \\ 
					MaskTrackRCNN~\cite{voigtlaender2019feelvos} & \checkmark & 12.3 & 59.9 & 26.2 & 9116 & 22.0 \\
					STEm-Seg~\cite{Athar_Mahadevan20ECCV}  & $\times$ & 12.2 & 58.2 & 25.4 & 8732 & 21.8 \\
					QDTrack-mots~\cite{qdtrack}  & \checkmark & 22.5 & 59.6 & 40.8 & 1340  & 22.4 \\
					QDTrack-mots-fix~\cite{qdtrack}  & \checkmark & 23.5 & 66.3 & 44.5 & 973 & 25.5 \\
					\midrule
					PCAN~\cite{pcan}  & \checkmark & 27.4 & 66.7 & 45.1 & 876 & 26.6 \\
					PCAN~\cite{pcan} + \textbf{BCNet} & \checkmark & \textbf{28.5} & \textbf{67.6} & \textbf{46.1} & \textbf{825} & \textbf{28.0} \\
					\bottomrule
			\end{tabular}}
		\label{tab:bdd}
		\vspace{-0.2in}
	\end{table}

\parsection{Qualitative Evaluation on COCO.}
Figure~\ref{fig:qualitative} shows 
qualitative comparison of CenterMask~\cite{lee2019centermask} and BCNet on images with overlapping objects using \textbf{FCOS detector}. In each ROI region, GCN-1 detects occluding regions while GCN-2 models the partially occluded instance by directly regressing the contours and masks. For example, BCNet decouples the occluding and occluded baseball players in similar clothes into GCN-1 and GCN-2 respectively, and detects the left leg missed by CenterMask. 
We also provide more qualitative results of our GCN-based BCNet compared to the Mask Scoring R-CNN~\cite{huang2019mask} on COCO {\it test-dev} set are shown in Figure~\ref{fig:qualitative_score}, both using ResNet-101-FPN and Faster R-CNN detector~\cite{ren2015faster}. Our proposed method is robust enough to deal with various occlusion cases, such as highly overlapping zebras and human hands. The contour and mask predictions by the two GCN layers for the occluder (GCN-1) and occludee (GCN-2) in the same ROI region also makes BCNet more explainable compared to previous methods.
In Figure~\ref{fig:qualitative_transformer}, we further show qualitative results comparison of transformer-based BCNet with single and bilayer transformer decoder, where our BCNet can even  handle well the highly occluded giraffe and motorcycle.

\parsection{Amodal results comparison on KINS} In Figure~\ref{fig:kins_compare}, we additionally provide qualitative \textbf{amodal} segmentation results comparison between Mask R-CNN + ASN module~\cite{qi2019amodal} and our BCNet on KINS~\cite{qi2019amodal} test set. Take the first case as an example, our BCNet infers more reasonable amodal car shape even when the front part of the car is heavily occluded by the standing woman.

\parsection{Evaluation on Video Instance Segmentation} For experiments on YTVIS, we replace the mask head of Mask Track R-CNN with our GCN-based BCNet. The results in Table~\ref{tab:vis} show an improvement of 2.1 AP. We also show one challenging qualitative results comparison in Figure~\ref{fig:vis}, where the overlapping regions between the two tandem skydivers are much better segmented by BCNet. For experiments on OVIS in Table~\ref{tab:ovis}, we adopt CMTrack RCNN~\cite{qi2021occluded} as the baseline, where BCNet achieves significant performance boost from 15.4 to 17.1, showing its efficacy of handling heavy occlusion in videos. Note that BCNet does not utilize temporal information while OVIS is a challenging video instance segmentation benchmark specifically designed to contain occluded video objects. 

\parsection{Evaluation on Multiple Object Tracking and Segmentation} For experiments on BDD100K MOTS, we augment the mask head of PCAN~\cite{pcan} with our GCN-based BCNet in Table~\ref{tab:bdd}, where MOTSA measures segmentation as well as tracking quality, and ID Switches
measure consistency in object identity. The quantitative results reveal an mAP advantage of 1.4 points, and mMOTSA gain over 1.0 points. The end-to-end training with new mask head also brings down ID Switches by 6\%  due to the improved instance features for association.  The advancements demonstrate that
our bilayer structure also generalizes to autonomous driving vehicles by providing more accurate segmentation masks.

\parsection{Limitation and Future Work} Although achieving large and consistent performance gain, we identify three design limitations for BCNet: 1) When dealing with unknown occluding objects of novel classes, the first GCN layer (transformer decoder) for detecting occluding objects may provide inaccurate occluder information for the second GCN layer (transformer decoder) to predict final occludee masks. This may cause BCNet to reduce to conventional instance segmentation models, outputting masks covering both the occluders and the occludee. For handling novel objects, one straightforward solution is to train BCNet in a class-agnostic manner as~\cite{vild}; 2) BCNet only focuses on the mask head design, thus the segmentation performance will be heavily influenced by the accuracy of the one/two-stage bounding box detectors; 3) BCNet is designed on single images which cannot utilize temporal cues in videos. Temporal information entails multiple views of the same dynamic moving objects for establishing correspondence. Further upgrading BCNet with temporal reasoning has the potential to further boost the  performance of detecting and segmenting  occluded video objects, a future research direction for pursuit.

\vspace{-0.2in}
\section{Conclusion}\label{sec:conclusion}

We propose BCNet, an effective mask prediction network for addressing instance segmentation in the presence of highly-overlapping objects in both image and video instance segmentation. BCNet achieves consistent  gains on overall performance using different backbones and one/two-stage object detectors in both the modal and amodal settings. We further explore the bilayer decoupling strategy on vision transformers (ViT) by representing instances in the image as separate occluder and occludee queries groups, and design the bilayer transformer decoder. With explicit occluder-occludee modeling, occluding and occluded instances are decoupled into two disjoint graph spaces, where the interaction between objects are explicitly considered. This effective approach will benefit future research in both occlusion handling and instance segmentation.

\ifCLASSOPTIONcaptionsoff
  \newpage
\fi




\vspace{-0.3in}
\bibliographystyle{IEEEtran}
\bibliography{egbib}
%



%
\vspace{-0.6in}
\begin{IEEEbiography}[{\includegraphics[width=1in,height=1.25in,clip,keepaspectratio]{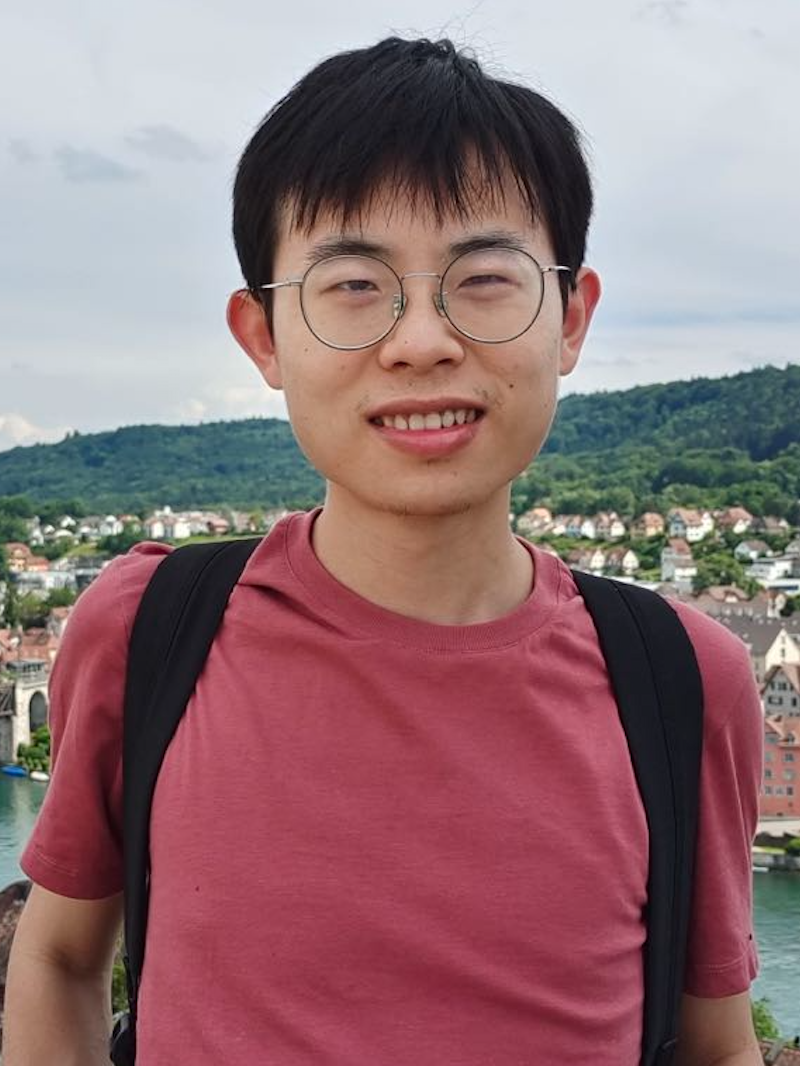}}]{Lei Ke} is a Ph.D. candidate in the Department of Computer Science and Engineering at the Hong Kong University of Science and Technology, advised by Chi-Keung Tang and Yu-Wing Tai. He is also a visiting scholar in the Computer Vision Laboratory of ETH Zürich since 2021. His research interests include image/video instance segmentation and object tracking. He received the BEng degree in Software Engineering from Wuhan University.
\end{IEEEbiography}

\vspace{-0.5in}
\begin{IEEEbiography}[{\includegraphics[width=1in,height=1.25in,clip,keepaspectratio]{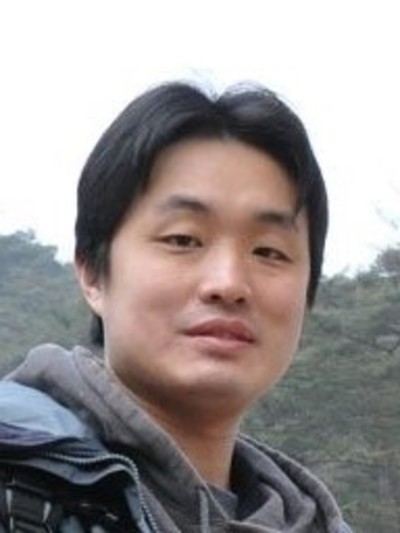}}]{Yu-Wing TAI} is a senior research director at Kuaishou Technology and an adjunct professor at CSE Department of HKUST. He received his BEng (First Class Honors) and MPhil degrees from the Department of Computer Science and Engineering, HKUST in 2003 and 2005 and PhD degree from the National University of Singapore in 2009. He was a research director of YouTu research lab of Tencent from January 2017 to April 2020. He was a principle research scientist of SenseTime Group Limited from September 2015 to December 2016. He was an associate professor at the KAIST from July 2009 to August 2015. He is an associate editor of IEEE Transactions on Pattern Analysis and Machine Intelligence (TPAMI). He regularly served as an area chair/technical program committee member of CVPR/ICCV/ECCV. His research interests include deep learning, computer vision and image/video processing. 
\end{IEEEbiography}

\vspace{-0.4in}
\begin{IEEEbiography}[{\includegraphics[width=1in,height=1.25in,clip,keepaspectratio]{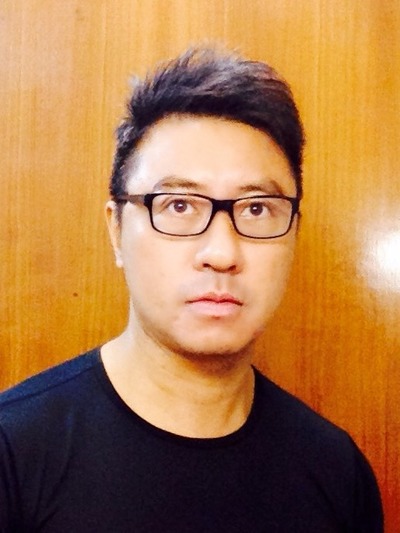}}]{Chi-Keung TANG} received the MSc and PhD degrees in Computer Science from the University of Southern California, Los Angeles, in 1999 and 2000, respectively.  Since 2000, he has been with the CSE Department at HKUST where he is currently a full professor.  He was on sabbatical at the University of California, Los Angeles, in 2008.  He was an adjunct researcher at the Visual Computing Group of Microsoft Research Asia.  His research areas are computer vision, computer graphics and machine learning. He was an associate editor of IEEE Transactions on Pattern Analysis and Machine Intelligence (TPAMI) and was on the editorial board of International Journal of Computer Vision (IJCV). He served as an area chair for ACCV 2006, ICCV 2007, ICCV 2009, ICCV 2011, ICCV 2015, ICCV 2017, ICCV 2019, ECCV 2020, CVPR2021, ICCV 2021, ECCV 2022, and as a technical papers committee member for the inaugural SIGGRAPH Asia 2008, SIGGRAPH 2011, SIGGRAPH Asia 2011, SIGGRAPH 2012, SIGGRAPH Asia 2012, SIGGRAPH Asia 2014 and SIGGRAPH Asia 2015.  He is a Fellow of the IEEE Computer Society, and has served on the IEEE Fellow Review Committee.
\end{IEEEbiography}




\end{document}